\documentclass[11pt]{article}
\usepackage{amsmath}
\usepackage{times}
\usepackage{graphicx}
\usepackage{subfigure}
\usepackage{color}
\usepackage{multirow}
\usepackage{rotating}
\usepackage{bbm}
\usepackage{latexsym}

\renewcommand{\baselinestretch}{1.4}
\renewcommand{\thesubsubsection}{\arabic{section}.\arabic{subsubsection}}

\newcommand{\captionfonts}{\normalsize}

\makeatletter
\long\def\@makecaption#1#2{%
  \vskip\abovecaptionskip
  \sbox\@tempboxa{{\captionfonts #1: #2}}%
  \ifdim \wd\@tempboxa >\hsize
    {\captionfonts #1: #2\par}
  \else
    \hbox to\hsize{\hfil\box\@tempboxa\hfil}%
  \fi
  \vskip\belowcaptionskip}
\makeatother

\begin{document}
\ \vspace{20mm}\\
{\LARGE When Not to Classify: Anomaly Detection of Attacks (ADA) on DNN Classifiers at Test Time\footnote{This work was
supported in part by a gift from Cisco and a grant from the DDDAS program at AFOSR.}}

\ \\
{\bf \large David J. Miller}\\
{djmiller@engr.psu.edu\\
    The Pennsylvania State University, School of EECS, University Park, PA 16802 USA\\}
{\bf \large Yujia Wang}\\
{upcheers@gmail.com}\\
    {The Pennsylvania State University, Department of Electrical Engineering, University Park, PA 16802 USA\\}
{\bf \large George Kesidis}\\
{gik2@psu.edu\\
    The Pennsylvania State University, School of EECS, University Park, PA 16802 USA\\}

\thispagestyle{empty}
\markboth{}{NC instructions}
\ \vspace{-0mm}\\
%
\begin{center} {\bf Abstract} \end{center}
A significant threat to the recent, wide deployment of machine learning-based systems, including
deep neural networks (DNNs),   
is {\it adversarial learning attacks}.
The main focus here is on evasion attacks against (DNN-based)
classifiers at test time.
While much work has focused on devising attacks that make perturbations
to a test pattern (e.g., an image) which are human-imperceptible and yet still induce
a change in the classifier's decision, until recently there has been relative paucity of work in defending
against such attacks. 
Some works robustify the classifier to make correct decisions on perturbed
patterns.  This is an important objective for some applications involving evasion attacks and for ``natural adversary'' scenarios.  However,
we analyze the possible evasion attack mechanisms and show that, in some important cases, when the image has been attacked, correctly classifying it has no utility:
i) when the image to be attacked is (even arbitrarily) selected from the attacker's cache; ii) when the sole recipient of the classifier's decision
is the attacker.
Moreover, in some application domains and scenarios it is highly actionable to detect
the attack irrespective of correctly classifying in the face of it (with classification still performed if no attack is detected).
We hypothesize that, even
if human-imperceptible, adversarial perturbations are machine-detectable.  We propose
a purely unsupervised anomaly detector (AD) that, unlike previous works: i) 
models the {\it joint} density of a deep layer using highly suitable null hypothesis
density models (matched in particular to the non-negative support for RELU layers); ii)
exploits multiple DNN layers;
iii) 
leverages a ``source'' and ``destination'' class concept, source class uncertainty, the class confusion matrix, and DNN weight information in constructing
a novel decision statistic grounded in the                                  
Kullback-Leibler divergence.                                      
Tested on MNIST and CIFAR-10 image databases under three prominent attack strategies, our approach outperforms previous detection methods, achieving strong ROC AUC detection accuracy on two attacks and
better accuracy than recently reported for a variety of methods on the strongest (CW) attack.  
We also evaluate a fully white box attack on our system.  Finally, we evaluate other important measures such as classification accuracy 
versus detection rate and multiple performance measures versus attack strength. 
\section{Introduction}
We are in the midst of a great era in machine learning (ML), which has found broad applications ranging from military, industrial, medical, multimedia/Web, and scientific (including genomics) to even the political, social science, and legal arenas. However, as ML systems are ever more broadly deployed, they become ever more enticing targets, both for individual hackers as well as for government intelligence services, which may seek to ``break'' them.  Thus, adversarial learning has become a hot topic, with researchers from both the security and ML communities devising various types of attacks and also defenses against same.  Focusing on statistical classification, prominent attack types include: i) data poisoning attacks on training data, with the typical goal to degrade a learned classifier's accuracy, e.g. \cite{Tygar11},\cite{Tygar14},\cite{Xiao15}, but more recently to create ``backdoors'', {\it e.g.} \cite{Song}; ii) {\it reverse engineering} attacks, which seek to learn a non-public (black box/undisclosed) classifier's decision-making rule by making numerous (even random) queries to the classifier \cite{Reiter}, \cite{Papernot3}; and, perhaps most importantly, iii) test-time evasion attacks \cite{Biggio_seminal}, wherein test (operational) examples are perturbed human-imperceptibly, but in such a way that the classifier's decision is changed (and now disagrees with a consensus human decision), e.g. \cite{Biggio_seminal},\cite{Szegedy_seminal},
\cite{Papernot},\cite{Goodfellow},\cite{CW}.  Such attacks, creating {\it adversarial 
examples} \cite{Szegedy_seminal}, may e.g. cause an autonomous vehicle to fail to recognize a road sign, or an automated system to falsely target a civilian vehicle.
Early seminal work in this area includes \cite{Biggio_seminal} and \cite{Szegedy_seminal}.
\cite{Biggio_seminal} formalized the problem in terms of the attacker's knowledge and
proposed a gradient-based approach to perturb patterns so as to induce misclassifications, applied to SVM classification of images and PDF files.
There are some related approaches for finding decision boundaries in neural networks that
are much older, e.g. \cite{Hwang}, but which were not applied to create adversarial attacks.
\cite{Szegedy_seminal} considered deep neural networks, posed a similar optimization problem as 
\cite{Biggio_seminal},
seeking imperceptible image perturbations that induce misclassifications, showed that this approach
is highly successful in creating adversarial examples starting from any (or at any rate, most) legitimate images, and also 
showed that these adversarial examples seem to generalize well (remaining as adversarial examples
for other networks with different architectures and for networks trained on disjoint training sets). 

\subsection{Critiques on Attacks, Which Motivate an AD Defense Strategy}

Recently, in \cite{MLSP17}, the authors published a somewhat critical review of research in this area.
Amongst other concerns, the ones that are of particular relevance to the present paper are the following:
\begin{enumerate}
\item Some evasion attacks, such as \cite{Papernot}, essentially presume that the classifier is defenseless.  In
particular, while the perturbed patterns in \cite{Papernot} do induce changes to the classifier's decision,
they also manifest artifacts (e.g. salt and pepper noise)
that are quite visible in their published figures.  These artifacts should be relatively easy to
(automatically) detect in practice.  Thus, their attack should only be successful against a defenseless system.  Moreover, the attack is not truly meeting the requirement for success that the attack
be human-imperceptible.
\item The test-time attacks in \cite{Goodfellow},\cite{Papernot},\cite{CW} assume that
the classifier must decide
on one of the $K$ known categories -- there is no possibility to {\it reject} a sample.
Classification with a rejection option\footnote{This is routinely applied, for example, by speech recognition systems like Siri.}
can provide optimal (minimum risk) decision rules when informed by the relative costs of various kinds of classification
errors and the cost of sample rejection.  Moreover, rejecting an attack sample is, in certain scenarios,
the decision with the least consequences, {\it e.g.}
i) leaking the least information to an attacker ({\it e.g.} in the case of a reverse engineering 
attack \cite{Reiter}), or ii) triggering a conservative action instead of an aggressive action, 
for applications where class decisions entail heavily consequential downstream actions (medical 
actions based on a classifier's automated image pre-screening decision, actuation of a robot \cite{Melis}, financial transactions, etc.).  
The potential value of rejection as an option in classifier decisionmaking has long been recognized, {\it e.g.}
\cite{Duda}.
Moreover, in security-sensitive settings, where the stakes for test-time attacks are the highest, the problem is often {\it not} classification per
se, but rather
{\it authentication}\footnote{As of 2015, there is a {\it speaker} recognition option as part of Siri.  Also,
authentication often involves simple multifactor criteria, {\it e.g.} CAPTCHAs.  Moreover, limited
privileges may be enforced, {\it e.g.} Siri cannot be used to enter data 
(including
passwords) into Web pages.}.
In a well-designed biometric authentication system, if there is any significant decision uncertainty (or atypicality) associated with the
presented pattern, the system will reject it (e.g., deny access to an individual).  In doing so, one is essentially
deciding that the given test pattern, though ``closest'' to a particular (authenticated/known) class, amongst
all such classes, is ``too anomalous'' relative to the typical patterns seen from that class -- moreover, one is weighing the cost of false
positives (invalid accesses) much higher than that of false rejections (invalid access denials).
\item \cite{Papernot},\cite{Goodfellow}, and \cite{CW} (strongly) assumed that the classifier structure and its parameter values are known to the attacker.  This is actually referred to as the
``zero knowledge'' case in \cite{Wagner17} because, while possessing full knowledge of the classifier, the attacker does
not possess knowledge of any defense ({\it i.e.}, attack detector) that may be in play.  Addressing the case where the classifier
is {\it not} (initially) fully known to the attacker, 
recent work has proposed techniques to reverse-engineer
a (black box) classifier without necessarily even knowing its structure or classifier type.  In \cite{Reiter}, the
authors consider black box machine learning services, offered by companies such as
Google, where, for a given (presumably big data, big model) domain, a user pays for class
decisions on individual
samples (queries) submitted to the ML service.  \cite{Reiter} demonstrates that, with a relatively modest number of queries (perhaps as many as ten thousand or more), one can learn
a classifier on the given domain that closely mimics the black box ML service decisions.  Once
the black box has been reverse-engineered, the attacker need no longer subscribe to the
ML service.
Moreover, such reverse-engineering enables test-time evasion attacks by providing knowledge
of the classifier. 
One weakness of \cite{Reiter} is that it neither considers very large (feature
space) classification domains nor very large networks (deep neural networks (DNNs)) -- orders
of magnitude more queries may be needed to reverse-engineer a DNN on a large-scale domain.  However, a much more critical weakness of \cite{Reiter} stems from one of its (purported) greatest
advantages -- the reverse-engineering in \cite{Reiter} does not require {\it any} labeled training
samples from the domain\footnote{For certain sensitive domains, or ones where obtaining real examples is expensive, the user may in fact have no realistic means of obtaining a significant number of real data examples from the domain.  This is one main reason why the ML service is needed in the first place -- the company or its client are the (exclusive) owners of this (labeled, precious) data resource, on the given domain.}.  In fact, in \cite{Reiter}, the attacker's queries to the black box
are randomly drawn, e.g. uniformly, over the given feature space.  While such random querying
is demonstrated to achieve reverse-engineering, what was not recognized is that
this makes the attack easily detectable by the ML service -- randomly
selected query patterns
are very likely to be extreme outliers, of all the classes.  Each such query is thus
individually highly suspicious by itself.  Multiple queries
should be easily detected as jointly improbable under a null distribution (estimable from the training set defined over all the classes from the domain).  Even if the attacker employed bots, each of
which makes a small number of queries, each bot's random queries should also easily be detected
as anomalous, likely associated with a reverse-engineering attack.  
Recently, reverse engineering attacks
based on more realistic queries have been proposed \cite{Papernot3}, leveraging some legitimate
training data from the domain.  However, the techniques
proposed here may provide effective means to detect even these (more subtle) reverse
engineering queries.  While the focus here will be on detection of test-time evasion attacks, detection
of reverse engineering attacks may be considered in future work. 
\end{enumerate}

What is central to all three of these critiques is that the papers \cite{Papernot},\cite{Goodfellow},\cite{Reiter}, and others do not consider the potential of a (purely unsupervised) {\it anomaly detection} (AD) approach for defeating
the attack.  While the likely effectiveness of AD to defeat \cite{Reiter} and even \cite{Papernot} 
(with noticeable salt and pepper noise) is clear,
it is less clear such an approach will be effective against human-imperceptible 
(or, at any rate, less perceptible) attacks such as
\cite{Goodfellow} and \cite{CW}.  This will be experimentally assessed in the sequel. 

\subsection{Defenses Against Test-Time Attacks:  Robust Classification Versus Attack Detection}

Irrespective of whether AD or some other defense approach, 
until the past few years there has been
relative paucity of work on defenses against test-time evasion attacks.
The basic premise and objective taken in some papers
\cite{Biggio},\cite{Giles},\cite{Papernot2},\cite{Li_ICCV},\cite{IBM-Ireland}  
is to robustify the classifier so that a test-time pattern that is a perturbation of a pattern from class A is still assigned to class A by the classifier.  

Robust classification
in general is an important objective, especially in the face of ``natural'' adversaries --
pattern variations induced by additive noise, transformations (image rotation, translation, and
scaling), etc.  A common approach in training DNNs is data augmentation, with such pattern variants
added to the training set so that the trained classifier is as robust as possible to such
variations.  However, the works \cite{Biggio},\cite{Giles},\cite{Papernot2},\cite{Li_ICCV},\cite{IBM-Ireland}
do not focus on (or even consider) achieving robustness to natural adversaries, but rather to (test-time) adversarial
attacks.  These methods are all also exclusively experimentally evaluated against adversarial attacks, not against natural
adversaries.  This is even true of \cite{IBM-Ireland}, which does use data augmentation to make
the classifier robust.  However, their method is only evaluated in adversarial attack settings,
not natural adversary settings (or some combination of both).

\cite{Biggio}
modifies the support vector machine (SVM) training objective to ensure the learned weight vector is not sparse.
Thus, if the attacker corrupts some features,
other (unperturbed) features still contribute to decisionmaking.
However, \cite{Biggio} may fail if
only a few features are strongly class-discriminating.  
\cite{Li_ICCV} performs blurring of test images in order to destroy a possible attacker's perturbations.
\cite{Giles}
considers DNNs for digit recognition and malware detection.  They
randomly {\it nullify} (zero) input features, both during training and use/inference.
This may
``eliminate'' perturbed features.  There are, however, several shortcomings here.
First, for the malware domain, the features as defined in \cite{Giles} are {\it binary} $\in \{0,1\}$.  Thus, nullifying (zeroing) does not
necessarily alter a feature's original value (if it is zero).  We suggest
recoding the binary features to $\{-1,1\}$.  Now, nullifying (zeroing) always changes the
feature value.  This may improve the performance of the method in \cite{Giles}.
Second, there is a
significant tradeoff in \cite{Giles} between accuracy in correctly classifying attacked examples and accuracy of the classifier in the absence
of attacks.  As the nullification rate is increased, the frequency of defeating the attack increases, but accuracy in the absence of attack decreases. For the CIFAR-10 domain \cite{cifar10}, in the best case,
one fourth of attack examples still cause misclassifications -- with significant loss in accuracy absent the attack. \cite{Papernot2} on the other hand reports relatively small loss in accuracy in the absence of attacks, for their ``distillation'' defense
strategy.  For \cite{Li_ICCV}, we will investigate classification accuracy in the absence of attacks
in our experimental results in the sequel.
It is also shown in \cite{Boult} that some attacks are quite fragile and that even the process of image
capture (involving cropping, etc.) may defeat the attack -- image capture itself, assuming the attack
is on a physical object, not an already captured digital image, may give 
robustness against adversaries.

\paragraph{``Don't know" response:}
However, a fundamental limitation of the robust classification approach to test-time evasion
attacks taken {\it e.g.} in 
\cite{Giles},\cite{Biggio}, and \cite{Papernot2}
concerns semantics of inferences.  Consider digit recognition.
Evasion-resistance means a perturbed version of `3' is still classified as a `3'.
This {\it may} make sense if the perturbed digit
is still objectively recognizable (e.g. by a human being) as an instance of a `3'.
In such case, it {\it may} be desirable to robustly classify this pattern as a `3' (It also may {\it not}
be desirable, even in this case, to make such a decision --  the decision may have no legitimate
utility.  This will be explained in the sequel).
However, the perturbed example may no longer be unambiguously recognizable as a `3' --
recall the perturbed digit examples from \cite{Papernot}, with significant salt
and pepper noise and other artifacts.  For some of the published images in \cite{Papernot}, ``don't know'' may be the most reasonable answer.
We will demonstrate such ambiguities later, in section 3.

\paragraph{Evasion-attack detection:}
Moreover,
irrespective of whether the perturbed pattern is class-ambiguous, it may be
important,
operationally, to recognize the classifier is being subjected to an evasion attack,
irrespective of
correctly classifying in the face of the attack.
Once an attack is detected, preventive measures to defeat the attack may be taken -- e.g., blocking
the access of the attacker to the classifier.  Also, human intervention or human consultation on
a final decision or action can be invoked \cite{Metzen}.  Moreover,
actions that are typically made based on the classifier's decisions may either be preempted or
(conservatively) modified.  For example, for an autonomous vehicle, once an attack on its
image recognition system is detected,
the vehicle may take the following action sequence: 1) slow down, move to the side of the road, and stop;
2) await further instructions.
Similarly, a machine that is actuated based on recognized voice commands might be put into a ``sleep''
mode, under which it can do no damage.  Similar ``conservative'' actions might be taken after attack
detection in other application domains (involving financial transactions, medical treatment, etc.) --
in a medical diagnostic setting, with an automated classifier either assisting human diagnosis
or used for pre-screening, one should not try to make a diagnosis based upon an attacked or fabricated
X-ray or MRI image.  If
one has the means to detect an evasion attack, one should do so prior to
performing any diagnosis.  If an attack is detected, a new image scan should be taken, with
the diagnosis then made based on trusted image data.  In \cite{Giles},\cite{Biggio},\cite{Papernot2} and
other papers, it is presumed that correctly classifying attacked test patterns is the right objective, without
considering attack detection as a separate, important inference objective. 

\subsection{Analysis of Test-Time Attack Mechanisms}

\begin{figure}[h]\centering
\centering\includegraphics[width=4in]{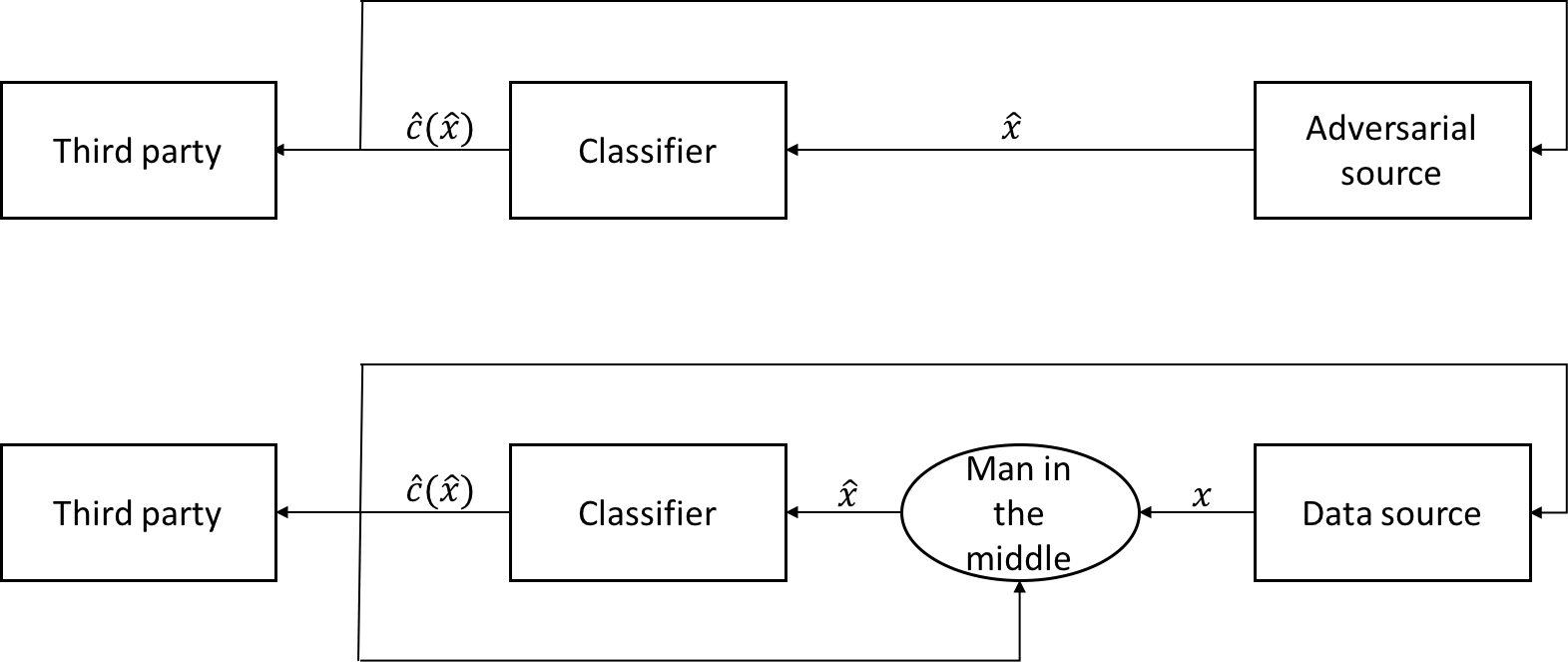}
\caption{The two possible test-time evasion attack mechanisms, with the attacker either an
adversarial source of images directly
or a man-in-the-middle, intercepting an image on its way to the classifier.  In the former mechanism, 
the image itself may be of no legitimate interest and the decision recipient is the adversary.  In the latter mechanism, the decision recipient could be either the honest image generator, another
intended recipient, or the adversary.}\label{fig:1}
\end{figure}

\begin{figure*}[!htb]
\includegraphics[width=1\columnwidth]{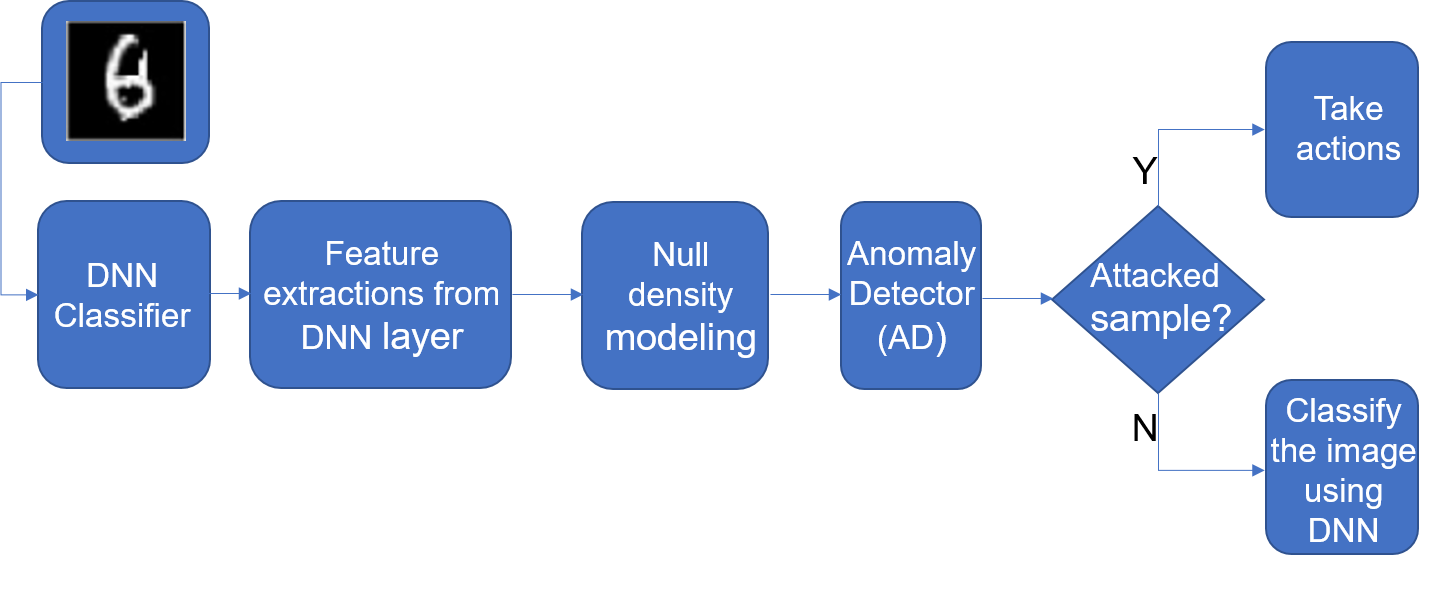}
\caption{Anomaly detection based on a decision statistic derived from multiple layers of a DNN.  Classification of the image is performed if no anomaly (attack) is detected.} \label{fig:2}
\end{figure*}

Beyond the above arguments that detection is 
extremely important, both operationally and in order to take risk-averse actions in some high stakes applications,
we also consider the problem formally.  The following analysis
is quite facile -- however,
it has not been given in any prior references of which we are aware.  Specifically, let us simply recognize
that there are only two mechanisms by which an attacked image is forwarded to a classifier (see Figure \ref{fig:1}).
In   
 one mechanism, there is
an honest generator of an image, but the image is then intercepted and perturbed by an adversary (essentially,
a ``man-in-the-middle'' attack\footnote{We assume no standard cryptographic
means are available to defend against man-in-the-middle attacks.}) 
before being forwarded to the classifier.
However, there is a second mechanism, wherein the adversary {\it is} the originator of the image.
Here, the adversary may have his own cache of labeled, legitimate examples from the domain.  He
selects one (even arbitrarily), perturbs it to induce a desired misclassification, and then 
sends it to the classifier\footnote{In both cases, if the attacker knows the classifier's decision rule, he can
make a perturbation that is guaranteed to induce a change in the classifier's decision, to a target class.}.  
Only for the first mechanism, where there is a legitimate image, forwarded by
an honest party who is interested in the classifier's decision, is it meaningful to try to correctly classify 
the attacked
image.
For the latter (adversarial source) mechanism, 
on an attacked image, the classifier's decisions are only being made for the possible benefit
of the adversary.
Moreover, even for the man-in-the-middle mechanism, it is not meaningful to make correct decisions
if they are only going to be intercepted by the adversary on their way back to the honest generator
(or to another intended decision recipient,
see Figure \ref{fig:1}).  We thus conclude that, if the image has been attacked,
it is only meaningful to seek to correctly classify when there is an honest generator {\it and}
when this honest generator (or another intended party) will be the (a) recipient of the classifier's decision.
While this may be a common scenario, we expect the adversarial source is also a common scenario.
However, even under the man-in-the-middle scenario, for the reasons articulated previously, attack detection
in high stakes domains (e.g., security settings) is important, irrespective of the importance of making correct decisions.
If no attack is detected, one can still make a best effort to correctly
classify the pattern.  This two-step process, with detection followed by classification when
no attack is detected, is the structure of our proposed system 
(see Figure \ref{fig:2}).
Finally, we note that a classifier may service multiple users, some legitimate and some
attackers.  Clearly, it is important to make correct decisions for legitimate users while at the same
time detecting attackers.  If the false detection rate of the system is set sufficiently low, our 
two-step process should be a practical solution in this multi-user setting.

There is one other important observation to make regarding Figure \ref{fig:1}.
In particular, note that the man-in-the-middle attack is actually not consistent with the common assumption made in the test-time
evasion attack literature \cite{Papernot},\cite{Goodfellow},\cite{CW},\cite{Wagner17}
that the attacker knows the ground-truth class label of the image he will perturb --
if the test image is forwarded to him/intercepted by him, he has no way of knowing its ground-truth label.  
Assuming the attacker knows the classifier, he can perturb the image to ensure there is a {\it change}
in the decision, but if the original image would have been incorrectly classified, the attacker's
perturbation could in fact lead to a correct classification, rather than a (desired) misclassification.
The exception is the situation where the attacker is actually a human being,
attacking a physical object, {\it e.g.} subtly 
modifying a road sign.  We consider this to be a man-in-the-middle attack where the attacker truly
{\it is} a human.  But if the attacker is an autonomous agent, attacking an intercepted digital image, the
ground-truth label of the image will not be known with certainty. 
Thus, our identification
of these fundamental attack mechanisms is further pointing out a potential problem with the assumptions being made in many adversarial learning attack (and defense) papers (if the attack is 
actually a man-in-the-middle).

\subsection{Recent Works on Detection of Test-Time Evasion Attacks}

Over the past several years, there has been great interest in detection of test-time evasion 
attacks, with a number of published works.
However, the problem has not been motivated based on the analysis we have just given.  One motivation
given in several works {\it e.g.} \cite{Wagner17} is that robust classification of attacked 
images is difficult, while detection is easier.  Certainly, given a robust classifier, one can
easily construct a detector.  Let the robust classifier's decision on image $X$ be $\hat{C}_{\rm r}(X)$.
One can also design a standard (non-robust) classifier, with decision $\hat{C}_{\rm s}(X)$.  One can
then make detections when the two classifiers disagree\footnote{We note, though, that using 
robust classifiers as detectors was neither proposed nor experimentally evaluated in
\cite{Giles},\cite{Biggio}, and \cite{Papernot2}
.}.  Likewise, considering methods like
\cite{Li_ICCV} which ``correct'' an image $X$ ({\it e.g.} by blurring it), producing a new image $X'$, one can build a single classifier and make an attack detection when $\hat{C}(X) \neq \hat{C}(X')$.
Even though a robust classifier can be readily used in these ways to make detections, that
does not mean this is a good approach for building a strong detector -- we will experimentally
evaluate ``detection via correction'' based on \cite{Li_ICCV} in the sequel, and show that our
direct design of a detector yields better detection and better classification performance 
(in the absence of attacks) than this strategy.

Recent published works have in fact directly tackled the attack detection problem.
Various detection approaches have been proposed, with a recent benchmark comparison study given
in \cite{Wagner17}, evaluating a number of methods against the CW attack \cite{CW}, which
was demonstrated in \cite{CW},\cite{Wagner17} to be more difficult to detect than earlier attack methods such as
\cite{Goodfellow} and \cite{Papernot}.
One strategy is to treat the detection problem as a supervised one, using labeled examples of ``known'' attacks.
The resulting binary classifier (attack vs. no attack) can then be experimentally evaluated both on
the known attacks and on unknown attacks.  Examples of such systems include \cite{Grosse} and
the supervised approach taken in \cite{AD} (we discuss \cite{AD} in detail, later).  However, \cite{Grosse} failed
to detect the CW attack on the CIFAR-10 image domain
\cite{Wagner17}.  \cite{AD}
similarly proved unsuccessful in detecting CW on CIFAR-10 \cite{Wagner17}.  
\cite{Li_ICCV} also treated the problem as supervised, applying a multi-stage classifier, with each
stage working 
on features derived from a deep layer of the trained DNN classifier.  A detection is made unless
all the stages decide the image is attack-free. \cite{Wagner17} demonstrates this detector performs
very poorly on CW applied both to the MNIST and CIFAR-10 domains. \cite{Metzen}, which
feeds a DNN classifier's deep layers as features to a second neural net supervised detector, is 
more successful.  However, the best results reported for this supervised method in detecting CW
on CIFAR-10 were 81\% true positive rate (TPR) at a false positive rate (FPR) of 28\% \cite{Wagner17}\footnote{The FPR is the number of non-attack test patterns that are detected as attack divided by the number of non-attack test patterns.  The TPR is the number of attack test patterns detected as attack divided by the numer of attack test patterns.}.
Our unsupervised AD will be shown to significantly exceed these results.  Moreover, \cite{Metzen}
reported that training on some attacks did not generalize well to other attacks (treated as unknown).
A related supervised approach is \cite{Safety-net}, which quantizes deep layer DNN features, feeding
the resulting codes as input to a support vector machine.  The authors argue that this quantization
makes their approach resilient even to white box attacks (where the attacker has full knowledge of
both the classifier and the detector) since gradients of their discriminant function are difficult
to compute (or even zero almost everywhere).  Again, this approach, being supervised, may not 
generalize well to unknown attacks.

Other approaches are unsupervised anomaly detectors, some based on explicit null hypothesis
density models for image features, with others taking a non-parametric approach, {\it i.e.}
without explicitly hypothesizing a stochastic model for non-attack data.   
One example of the latter is simply to make detections when decision ``confidence''
is below a given threshold, {\it e.g.} \cite{Melis}.  In particular, one can reject if the maximum
{\it a posteriori} class probability (produced by the classifier) is less than a given threshold.
Use of such confidence was shown to be effective for detection of a classifier's {\it misclassified} samples
in \cite{Hendrycks_misclass}, although effective detection of adversarial attacks may require more powerful
detectors.
Another non-parametric approach is based on principal component analysis (PCA) \cite{Hendrycks}.
However, \cite{Wagner17} found that, while successful on MNIST, this approach has esssentially
no power to distinguish attacks from non-attacks on CIFAR-10.  A more sophisticated, interesting non-parametric
detector is \cite{Magnet}.  Extending beyond PCA's linear representation, \cite{Magnet}
extracts nonlinear components via an auto-encoding neural network. 
\cite{Magnet} uses somewhat unconventional decisionmaking, {\it combining} classification
and detection -- images whose (auto-encoding based) reconstruction error is large (far from the
non-attack image manifold) are detected as
attacks.  Images whose reconstruction error is small are sought to be correctly classified.
However, given our earlier arguments, it may be desirable to make detections even when the
attack is subtle, with the attacked image lying close to the non-attack image manifold.  A primary concern with this 
detector is that it requires setting a number of hyperparameters (specifying the auto-encoding
network architecture, which in fact required different settings for MNIST and CIFAR-10, and softmax
``temperature'' variables).  Setting hyperparameters in an unsupervised AD setting is difficult.
If labeled examples of an attack are available, one can set hyperparameters to maximize a validation set
measure.  However, the attack is then no longer ``unknown'' and the detection method is actually
supervised.  \cite{Magnet} also incurs some degradation in classification accuracy in the 
absence of an attack -- from 90.6\% to 86.8\% on CIFAR-10 \cite{Magnet}.

One parametric detection approach is \cite{Openmax}.  This approach computes the distance
between a deep layer's class-conditional mean feature vector and the test image's feature vector
and then evaluates this, under the null hypothesis of no attack, using a Weibull distribution.
A few limitations of \cite{Openmax} are that: i) it does not model the {\it joint} density of a deep
layer -- such a model would exploit more information than just the (scalar) distance; ii) \cite{Openmax} is not truly
unsupervised -- several hyperparameters are set 
by maximizing a validation measure that requires 
labeled examples of the 
attack.  In our experiments, we will show that our purely unsupervised AD outperforms \cite{Openmax}
even though we (optimistically) allow their detector design to use more than 100 labeled examples of the 
attack to be detected.
More recently, \cite{AD} did propose a method based on a null hypothesis joint density
model of a deep layer feature vector.  To our knowledge, theirs is the first such approach,
upon which our detection methodology builds.  \cite{AD} used a kernel density estimator to
model the penultimate layer of a DNN.  However, ultimately, they put forward a {\it supervised}
method, learning to discriminate `attack' from `no attack', that leverages their density model to create the classifier's input features.  They ultimately settled on a supervised method because their
unsupervised detector did not achieve very good results.  In \cite{Wagner17}, it was found that
the unsupervised AD in \cite{AD} 
grossly fails on CW attacks on CIFAR-10 -- 80\% of the
time, attacked images had even {\it higher} likelihood under the null model in \cite{AD} than the original (unattacked)
images.
Our unsupervised AD, developed in the next section, is based on a number of novel innovations
beyond the basic null density modelling in \cite{AD} and will be shown to achieve substantially
better results than this detector.

\subsection{Contributions of This Work}

\begin{enumerate}
\item As already elucidated, we have identified not only scenarios and applications where attack detection
is important but also attack
mechanisms for which, in the presence of an attack, robust classification is only for the benefit of the attacker, while detection remains an important objective.
\item We develop a novel unsupervised AD that goes well beyond \cite{AD}, 
i)
modeling the {\it joint} density of a deep layer using highly suitable null hypothesis
density models (matched in particular to non-negative support for RELU layers); ii)
exploiting multiple DNN layers;
iii)
leveraging a ``source'' and ``destination'' class concept, source class uncertainty, the class confusion matrix, comprehensive {\it low-order} density modelling \cite{Qiu16}, and DNN weight information in constructing
a novel decision statistic grounded in the
Kullback-Leibler divergence.
\item We demonstrate state of the art results on published attacks, including CW \cite{CW},
compared against a number of benchmark detection methods and compared with the
results reported in \cite{Wagner17}.
\item We evaluate some performance measures of great interest that are sometimes not assessed
in prior works, including multiple performance measures as a function of attack strength and 
classification accuracy
versus the false positive rate of attack detection.
\item We develop a novel white box attack, extending the approach in \cite{Wagner17} to attack a system consisting of a detector and a classifier.  We evaluate
this attack against our proposed system.
\end{enumerate}
\section{Proposed Method: Anomaly Detection of Adversarial Attacks (ADA)}
\subsection{Defender's Knowledge, Goals, and Assumptions}
The defender is designed to detect test-time evasion attacks and to correctly
classify if no attack is detected (Figure \ref{fig:2}).  The defender assumes very little about
the attacker -- simply that the attacker 
may have full knowledge of the classifier being used and, in performing attacks,
will start from a legitimate image from the domain (from some class, $c_s$) that is correctly classified by
the classifier,
and seek to make human-imperceptible perturbations to 
the image causing the classifier's decision to change from $c_s$ to another class.
Human-imperceptibility connotes two senses:
i) a human being would classify the image to class $c_s$;
ii) the image perturbations should not be noticeable, {\it i.e.} they should not raise suspicion that there has been image tampering.
The defender does not know whether an attack is present
in a given image and if one is present, does not know whether the attack is targeted
(changing the decision to a specific targeted class $c_d$) or indiscriminate \cite{Tygar}.
The defender is unaware whether the
attacker has any knowledge of the detector (even {\it if} the attacker knows a detector is being used).  The defender does not specifically {\it rely} on the detector being unknown to the attacker
(security by obscurity \cite{Biggio_KDE}) -- even if the detector is known to the attacker,
perturbing to defeat both the classifier and the detector may be difficult and/or
may fail the human-imperceptibility attack requirement.  However, the defender also recognizes
the difficulty in devising a defense to defeat an attack that is truly white box with respect to
both the classifier and detector \cite{Magnet}.
The defender's goals are to achieve: i) high attack true detection rate (TDR) at a low
attack false positive rate (FPR) and ii) high classification accuracy conditioned on no attack
detected (with the detector set to achieve a low FPR).
The defender's hypothesis is that, to remain human-imperceptible but still to induce 
misclassifications on a DNN, the image perturbation is likely to result in a detectable
signature in some of the DNN's (deep) layers.   
Our ADA approach is a novel detection framework built
around this hypothesis.
No strong restrictions on the defender's resources (off-line training computation and memory,
operational computation and memory) are assumed.
\subsection{Attacker's Knowledge, Goals, and Assumptions}
In most of our experiments (section 3), we evaluate the published attacks
in \cite{Papernot},\cite{Goodfellow}, and \cite{CW}.
These exploratory \cite{Tygar}, test-time evasion attacks are ``white box'' with respect
to the attacker's knowledge of the classifier \cite{Biggio_wild} but ``zero knowledge'' (black box) with 
respect to the detector \cite{Wagner17}.  That is, they exploit full knowledge of the classifier
(both to evaluate the classifier's decisions and to compute gradients with respect to
the classifier's discriminant function, for creating perturbed images). They also exploit a
``cache'' of legitimate image examples from the classification domain of interest, with true class
labels known.  
Thus, the attacker can perturb one of these images to induce a change to the classifier's decision
and can verify that the change is a misclassification, based on knowledge of the true class
label\footnote{Note that this assumption is inconsistent with a ``man-in-the-middle'' attack,
where the image to be classified is intercepted by the attacker (with its true class of origin
unknown to the attacker).}.
The attacks in                                                                         
\cite{Papernot}, \cite{Goodfellow}, and \cite{CW} are {\it targeted}, seeking to change
an image with true label $c_s$ (that is correctly classified) into an image that is misclassified
to a target class $c_d$.
A successfully attacked image means that a targeted misclassification was achieved 
and that the image perturbation is human-imperceptible (as defined in the previous subsection). 
A successful attack (imperceptible misclassification) may still be detected by an AD.
The attacker's cache could consist of the training images used to learn the classifier, but it also could be a separate data resource.  
The cache needs to be ``rich'' in that it possesses
(a sufficient number of) examples from all classes -- in this way, the attacker can produce targeted
misclassifications from any starting class to any destination class.  Further, it can produce
many such attack examples (by perturbing starting from many different images).   
The attacker's goal is to design an attack that achieves as high a success rate as possible.

In addition to evaluating these attacks that have no knowledge of the detector, in one experiment we also
evaluate a complete {\it white box} attack, wherein the attacker possesses full knowledge
of both the classifier {\it and} the detector.
In this case, a successful attack is one which is imperceptible, induces a misclassification, {\it and}
is not detected.
\subsection{Notation and Setup}
Consider a ``raw'' feature vector $\underline{x} \in {\cal R}^d$, which could e.g. represent a (scanned)
array of gray scale values comprising a digital image.
Consider an $L$-layer DNN.
Let $P_{\rm DNN}[C=c | \underline{x}], c=1,\ldots, K$
be the {\it a posteriori} probability that $\underline{x}$ originates from class $c$, amongst the categories
in a classification problem with $K$ known categories.  Without loss of generality we represent these categories
by the
integers $\{1,2,\ldots,K\}$.  There is a {\it labeled training set} ${\cal X} = \{{\cal X}^{(c)}, c=1,\ldots,K\}$,
where ${\cal X}^{(c)} = \{\underline{x}_1^{(c)},\underline{x}_2^{(c)},\ldots,\underline{x}_{N_c}^{(c)}\}$
are the labeled training samples from class $c$.  We have two purposes for this training set.  First,
it is used to learn the DNN posterior model, $P_{\rm DNN}[C | \underline{x}]$ (via a suitable DNN training
method).  Second, suppose that $\underline{z} \in {\cal R}^{d(l)}$ is the output vector for layer $l$ of the DNN,
$l \in \{2,3,\ldots,L-1\}$,  when
$\underline{x} \in {\cal R}^d$ is the input to the DNN -- layer $l$ could be sigmoidal, an RELU, or even a
max-pooling layer of the DNN.  Then, by feeding each of the training examples from class $c$,
$\underline{x}_i^{(c)}$, into the
(already trained) DNN and extracting the layer $l$ output vector for each such example, we can create a layer $l$
derived feature vector training set conditioned on class $c$ (with explicit notational dependence on $l$ omitted
for simplicity), {\it i.e.} ${\cal Z}^{(c)} = \{\underline{z}_1^{(c)},\underline{z}_2^{(c)},\ldots,\underline{z}_{N_c}^{(c)}\}, c=1,\ldots, K$.
For each such derived training set ${\cal Z}^{(c)}$, representative of class $c$, one can learn the class-conditional
joint density, assuming a particular parametric density form and performing suitable model learning ({\it e.g.},
maximum likelihood estimation, coupled with model-order selection techniques such as Bayesian Information
Criterion \cite{Schwarz}, to estimate the model structure and ``order'' ({\it e.g.}, the number of components,
in the case of a mixture density)).  Denote the resulting learned class-conditional densities for a particular
layer of the DNN classifier by $f_{\underline{Z}|c}(\underline{z} | c), c=1,\ldots, K$.  These densities,
over all layers being modelled,  together constitute a
``null hypothesis model'' --
where the null hypothesis is that a test vector $\underline{z} \in {\cal R}^{d(l)}$
is the result of feeding in an {\it unperturbed} image $\underline{x}$ from one of the $K$ categories into the DNN
and extracting the $l$-th layer output of the DNN, $\forall l$.
The alternative hypothesis, accordingly, is that $\underline{z}$ is the result of feeding an attacked
(perturbed) image, call it $\underline{x}'$, into the DNN.
\subsection{The Method from \cite{AD}}
The AD proposed in \cite{AD} consists of the following operations, given a test pattern (image), $\underline{x}$:
\noindent
\begin{enumerate}
\item Determine the maximum {\it a posteriori} class under the DNN model:
\\
$c^{\ast} = \arg\max_{c \in \{1,\ldots,K\}} P[C=c | \underline{x}]$.
\item Compute $\underline{z} = \underline{g}^{(l)}(\underline{x})$, where $\underline{g}^{(l)}()$ is the
function whose input is $\underline{x}$ and whose output is the layer $l$ output of the DNN.
\item Evaluate $f_{\underline{Z}|c^{\ast}}(\underline{z} | c^{\ast})$ and declare an attack detection if this
value is below a preset threshold.
\end{enumerate}

Beyond the above-described procedure, full specification of the method in \cite{AD} requires the choice of the
layer, $l$, and the density function ``family'' $f_{\underline{Z}|c^{\ast}}()$.  \cite{AD}
used the penultimate layer of the DNN (the layer immediately preceding the decision layer, $l=L-1$) and chose a
simple, Gaussian kernel-based density estimator.  We will discuss these choices further below.  However,
without even considering these choices, we propose an improved detector, which fundamentally
exploits more information about a possible anomaly, to strong effect (as demonstrated by our results),
compared to the procedure from \cite{AD}.

\subsection{The Basic Proposed ADA Method}
\begin{figure}[h]
\centering\includegraphics[width=0.3\columnwidth]{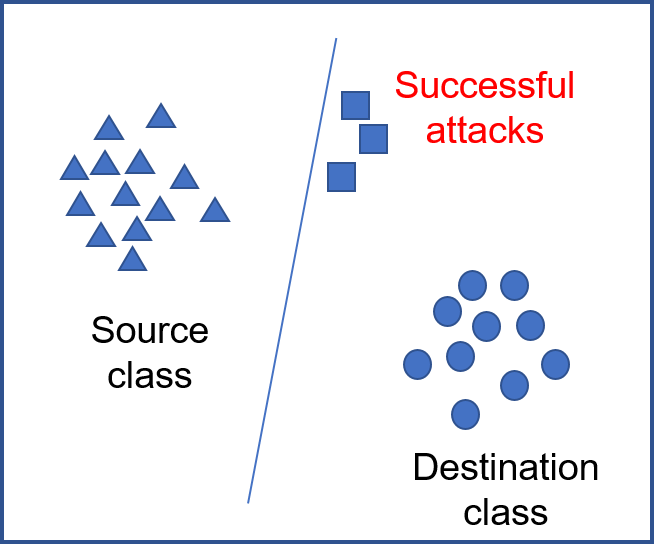}
\caption{Illustrative attack examples in the plane -- such examples may have relatively high likelihood under the source class and low likelihood under the destination
class (the class decided by the classifier on the attacked image).}
\label{fig:3}
\end{figure}

Consider a successful attack example -- one which was obtained by starting from a ``clean''
example $\underline{x}$ from an (unknown)
source class $c_s$ and then perturbing it until the DNN's decision on this perturbed example (now $\underline{x'}$)
is no longer $c_s$, but is now $c_d \neq c_s$ (the ``destination'' class).  The premise behind the approach
in \cite{AD} is that a test pattern $\underline{z}$ which results from feeding an {\it attacked} version of $\underline{x}$, not $\underline{x}$ itself, into the DNN,
will have atypically low likelihood under the density model for the DNN-predicted class $c_d = c^{\ast}$.
While we expect this may be true, if the perturbation of $\underline{x}$ is not very large
(consistent with its human-imperceptibility), we might {\it also} expect that $\underline{z}$ will exhibit
{\it too much typicality} (too high a likelihood) under some class other than $c^{\ast}$, {\it i.e.}
under the source category, $c_s$.  It does not matter that the source category is unknown.  We can
simply determine our best estimate of this category as: ${\hat c}_s = \arg\max_{c \in \{1,\ldots,K\} - c^{\ast}}
f_{\underline{Z}|c}(\underline{z}|c)$, with the associated ``typicality'' $\max_{c \in \{1,\ldots,K\} - c^{\ast}}
f_{\underline{Z}|c}(\underline{z}|c)$.
Accordingly, we hypothesize that attack patterns should be {\it both} ``too atypical'' under $c^{\ast}$
and ``too typical'' under ${\hat c}_s$.  
This is illustrated in Figure \ref{fig:3}.

While this may seem to entail an unwieldy detection strategy that
requires use of two detection thresholds, we instead propose a single, theoretically-grounded
decision statistic that captures both requirements (jointly assessing the ``atypicality'' with 
respect to $c^\ast$ and the ``typicality'' with respect to ${\hat c}_s$).
Specifically, define a two-class posterior evaluated with respect to the (density-based) null model:
$P \equiv [P_{c^{\ast}},P_{\hat{c}_s}] = [p_0 f_{\underline{Z}|c^{\ast}}(\underline{z}|c^{\ast}),p_0 f_{\underline{Z}|{\hat c}_s}(\underline{z}|{\hat c}_s)]$,
where $p_0$ gives the proper normalization: $p_0 = (f_{\underline{Z}|c^{\ast}}(\underline{z}|c^{\ast}) + f_{\underline{Z}|{\hat c}_s}(\underline{z}|{\hat c}_s))^{-1}$.
Likewise, define the corresponding two-class posterior evaluated via the DNN:
$Q \equiv [Q_{c^{\ast}},Q_{\hat{c}_s}] = [q_0 P_{\rm DNN}[C= c^{\ast} | \underline{x}], q_0 P_{\rm DNN}[C = {\hat c}_s | \underline{x}]]$,
where $q_0 = (P_{\rm DNN}[C= c^{\ast} | \underline{x}] + P_{\rm DNN}[C= {\hat c}_s | \underline{x}])^{-1}$.
Both deviations (``too atypical'' and ``too typical'') are captured in the Kullback-Leibler divergence 
decision statistic: $D_{\rm KL}(P || Q) = \sum\limits_{c \in \{c^{\ast},\hat{c}_s\}} P_c\log(\frac{P_c}{Q_c})$, {\it i.e.} we declare a detection when this statistic exceeds a
preset threshold value.
Basic pseucodode for the resulting ADA procedure is given below:

\noindent
{\bf Model Learning:} 
\begin{enumerate}
\item Learn the DNN classifier $P_{\rm DNN}[C | \underline{x}]$ using the given training set.
\item Estimate the class-conditional null densities $f_{\underline{Z}|c}()$, for all classes, $c$,
for the specified layer, $l$, where $\underline{Z} = \underline{g}_l(\underline{X})$.
\end{enumerate}

\noindent
{\bf ADA Inference:}
\begin{enumerate}
\item Given a test image $\underline{x}$, compute the two-class posterior pmfs $P$ and $Q$ specified above.
\item For the given 
detection threshold $\tau$, declare a detection if $D_{\rm KL}(P || Q) > \tau$.
\item If no detection is made, classify $\underline{x}$ using the maximum {\it a posteriori} rule applied to the
DNN posterior $P_{\rm DNN}[C= c^{\ast} | \underline{x}]$.
\end{enumerate}
 
We note that KL divergence has been used in many works as an AD 
statistic, {\it e.g.} \cite{Afgani},\cite{Xu_KL},\cite{Najah}.
In \cite{Xu_KL}, KL is actually used in a supervised setting (to test between normal
and abnormal class pmfs, with the abnormal class pmf estimated based on supervising information).  In \cite{Afgani},
detections are made over time windows, with a pmf under the putative alternative hypothesis estimated using observations in the time window under consideration.
Our application of KL divergence is within a purely unsupervised AD setting 
(labeled anomaly (attack) examples are not available for alternative model learning),
unlike \cite{Xu_KL}.  Moreover, our method does not require on-line or time-window based estimation of
a putative alternative hypothesis pmf, unlike \cite{Afgani}. 
Finally, unlike these past works, our use of KL does not require {\it explicitly} forming an alternative
hypothesis pmf --
our KL is evaluated over pmfs defined on the classes in a supervised classification problem, with
the class pmf evaluated in two ways: 1) using the DNN classifier's posterior; 2) using class-conditional null distributions derived from the DNN's deep feature layers.
This allows us to have a single (KL-based) decision statistic that
simultaneously assesses being ``too atypical'' with respect to one class (the attacker's destination class)
and ``too typical'' with respect to another class (the attacker's putative source class).  While we are not claiming high novelty of this approach,
we have not seen KL used before for AD in the fashion we propose here.

Note that one of (several, possible) interpretations for the (asymmetric) roles of $P$ and $Q$
in $D_{\rm KL}(P || Q)$ is that $P$ represents the ``true'' posterior, with $Q$ an alternative model posterior
whose ``agreement'' with the true is what we would like to assess.  We place the null model posterior estimate
in the ``true'' position because we believe it better reflects actual class uncertainties in a test pattern
$\underline{x}$ than does the DNN posterior -- there is some empirical evidence for this in our experiments.
Moreover, in our experiments we have found that this choice leads to better AD prediction power than what
one gets by exchanging the roles of $P$ and $Q$ (and also modestly better than what one obtains using
a symmetric version of KL divergence).

We emphasize that our hypothesis that an attacked pattern will exhibit both ``too high atypicality'' (with respect to $c^{\ast}$) and
``too high typicality'' (with respect to ${\hat c}_s$) is not based on any strong assumptions about the
attacker.
It merely assumes the attack is seeking to be imperceptible to a human being.  To achieve this,
the attacked example should be definitively recognizable by a human being as a legitimate ({\it visually} artifact-free) instance of the source class 
($c_s$).  This constraint, in conjunction with the attacked example's misclassification by the classifier, 
may necessitate that the test pattern will exhibit unusually high
likelihood for a category ($c_s$) other than that decided by the DNN.  At the same time, because the
(successful) attack example should not appear to a human to belong to $c^{\ast}$, this may necessitate the test
pattern will exhibit unusually {\it low} likelihood under $c^{\ast}$.

Note that it is not so straightforward to ``bin'' our proposed detection principle.  Specifically, 
\cite{Lecapitaine}
refers to ``distance-based'' rejection
of samples that are outliers of known categories
and ``ambiguity-based'' rejection of samples for which there 
is too much ambiguity to decide between several competing (inlier) classes.
Our approach is {\it neither} purely distance-based nor
ambiguity-based -- in making a detection, the sample is being assessed to have ``too great distance'' with
respect to the destination class ($c^{\ast}$) and simultaneously to have class ambiguity, based on its likelihood
under a source class ($\hat{c}_s$) being relatively large whereas its classifier discriminant output is largest
for a different (the destination) class.
\subsection{ADA Improvements}
\begin{enumerate}
\item {\it Layer and Null Model Choices}: \cite{AD} chose $l=L-1$, the penultimate layer of the DNN, and used a Gaussian kernel-based density estimator.  We investigate Gaussian mixture models (GMMs) for layers with full (positive
and negative) support. As a novel contribution, we also propose
 log-normal multivariate mixture density modelling for RELU activation layers to reflect the fact that the RELU 
has unbounded {\it non-negative} support.  
This involves
log-transforming all features in the layer and then applying Gaussian mixture modelling (GMM)
to these transformed features.  For an RELU, applying the log() yields deep layer features with {\it full} (positive
and negative) support (consistent with the full support of the GMM).  Experimentally, we have found this yields performance gains
over Gaussian kernel and more general GMM joint densities. By contrast, \cite{AD} 
directly modelled non-negative deep layer features using Gaussian kernel densities (a special
case of GMMs),
which have full (positive and negative) support -- this suboptimally gives probability support to negative feature values, even though the RELU features are non-negative;
layers with full support (e.g. a hyberbolic tangent) can be directly modeled via GMMs.
We also investigate
{\it several} different layers, $l \in \{2,3,\ldots,L-1\}$, for extraction of feature vectors.  
\item {\it Maximizing KL over different layers}:
Rather than
restricting to a {\it single} layer, for a given test image $\underline{x}$ one can measure the KL divergence at
{\it multiple} layers and choose, as the decision statistic, the {\it maximum} KL divergence over the different
layers.  This may enhance detection performance, as anomalous signatures may not always prominently manifest
in the same ({\it e.g.,} penultimate) layer.  We refer to this approach as ADA-maxKL.  We will report
how frequently different layers (close to the image input, or close to the DNN output) are the ``winner'' in 
ADA-maxKL.
\item {\it Considering All Classes}:
Instead of just considering $c_d$ and $c_s$, it is also possible to form probability vectors for $P$ and $Q$
over {\it all} classes.  While our detector's underlying hypothesis suggests that most of the anomalous signature may
manifest with respect to $c_d$ and $c_s$, more information may be exploited by considering all classes (especially if some classes are similar to each other).
\item {\it Exploiting uncertainty about $c_s$ and knowledge of class confusion}:
ADA as defined so far makes a hard decision estimate of the source class, $c_s$.  Alternatively, we can estimate
the probability that $C_s=c$ via 
$${\sf P}[C_s=c] = \frac{f_{\underline{Z}|c}(\underline{z}|c)}{\sum\limits_{c' \in \{1,\ldots, K\} - c^\ast} f_{\underline{Z}|c'}(\underline{z}|c')}, ~c \neq c^\ast.$$
Furthermore, suppose we have knowledge of the classifier's confusion matrix $[{\sf P}[C^{\ast}=i | C=j]]$, reflecting
normal class confusion in the absence of an attack (obtained
{\it e.g.} from a validation set).  Then, a class pair $(c_s,c_d)$ with very small confusion probability
${\sf P}[C^{\ast}=c_d | C= c_s]$ is much more likely associated with an attack than a pair with high 
class confusion.  Accordingly, we suggest to weight the KL divergence by $\frac{1}{{\sf P}[c_d | c_s]}$\footnote{We replace zero probabilities in the confusion matrix by small values and renormalize so that it remains a pmf, conditioned
on each class $c_s$.}.  This weighting increases the contribution to the decision statistic for pairs unlikely to occur
due to normal (non-attack) classifier confusion.
Combining {\it both} these ideas, we construct the Average, Weighted ADA (AW-ADA) statistic for a given layer
as: 
$$\sum\limits_{c \neq c^{\ast}} {\sf P}[C_s=c]\frac{D_{KL}(P^{(c)} || Q^{(c)})}{{\sf P}[C^\ast=c^{\ast} | C=c]},$$
where $P^{(c)} = [p_0 P[c^{\ast} | \cdot],p_0 P[c | \cdot]]$ and $Q^{(c)} = [q_0 P_{\rm DNN}[c^{\ast} | \cdot], q_0 P_{\rm DNN}[c | \cdot]$.  Moreover, this can be evaluated for different layers, with the 
decision statistic its {\it maximum} over all considered layers.  The resulting approach is dubbed
AW-ADA-maxKL. As will be seen, this approach achieves significant
improvement in detection accuracy over the basic ADA method and over \cite{AD}, especially on the CIFAR-10 data set.
\item {\it A ``Local'' Version of AWA-ADA-maxKL}:
Instead of modelling the {\it joint}
feaure vector for a layer, we can alternatively comprehensively model low-order feature
{\it collections} -- {\it in particular}, all feature pairs (each pair via a log-normal mixture density {\it e.g.} if the activation function is an RELU).
This is a rich ``low order'' feature representation -- for example, for a layer $l$ with 400
features, there are $N_l =$ (400 choose 2) feature pairs that will be individually modelled.
For
{\it each} such feature pair, $(Z_i,Z_j)$, for each layer being modeled, one can separately form the AW-ADA 
KL-based statistic.
Moreover, each such feature pair's AW-ADA statistic can be {\it weighted} based on the magnitude of the DNN weights from these features ($Z_i$ and $Z_j$) to the {\it next} layer of the DNN (higher magnitude DNN weights indicate the feature pair 
$(Z_i,Z_j)$ is important and that its AW-ADA statistic should be given stronger influence than pairs with lower magnitude DNN weights).  Accordingly, for each layer we form a weighted aggregation of all low-order
AW-ADA statistics, 
expressed for layer $l$ as:
$${\rm L-AWA-ADA}^{(l)} = \frac{1}{N_l}\sum\limits_{\forall(i,j)} \beta_i \beta_j \sum\limits_{c \neq c^{\ast}} {\sf P}_{ij}[C_s=c]\frac{D_{KL}(P_{ij}^{(c)} || Q^{(c)})}{{\sf P}[C^\ast=c^{\ast} | C=c]}.$$
We then choose, as the decision statistic, the maximum of this quantity over all layers $l$ being modelled.
Here, $P_{ij}^{(c)} = [p_0 P[c^{\ast} | (z_i,z_j)],p_0 P[c | (z_i,z_j)]]$, which depends on the feature pair
$(z_i,z_j)$, while $Q^{(c)} = [q_0 P_{\rm DNN}[c^{\ast} | \underline{x}], q_0 P_{\rm DNN}[c | \underline{x}]]$
and is not a function of $(i,j)$.
Also, here,
$\beta_i$ is the sum of the magnitudes of the DNN weights that conduct from feature $i$ in layer $l$ 
to all neurons in the {\it next} layer, $l+1$, normalized by the {\it maximum} such sum over {\it all} features in layer $l$.
Note also the normalization by the number of feature pairs in layer $l$, $N_l$.  This is done
so that, in {\it maximizing}  L-AWA-ADA$^{(l)}$ over all layers $l$, there is a fair comparison of the statistics produced by each of the layers (which may have different numbers of features).
Finally, note that we have omitted dependence on the layer, $l$, in the LW-AWA-ADA$^{(l)}$ expression (except in the $N_l$ term)
for concision of expression.

We dub the resulting method L-AWA-ADA-maxKL, `L' for ``local''.
The motivation for this approach is that atypicalities may manifest
on a {\it very small} subset of the features in a layer -- if one null-models the {\it joint}
feature vector for a layer, atypicalities in just a few features may yield only very weak assessed joint feature atypicality.  On the other hand, a high degree of atypicality will be assessed for
a low-order feature collection that contains the strongly atypical features.  This approach is inspired by
 the active learning based class discovery framework in \cite{Qiu16}, where it was found that comprehensively modeling {\it low-order} null densities leads to excellent detection
of anomalies with small ``footprints'' (only a few abnormal features, amongst many measured
features).  This approach will be seen to be our preferred ADA method, giving the overall best results
across all tested attacks. 
\end{enumerate}

\section{Experimental Results}
We have evaluated our proposed ADA detection method and its variants in comparison with a variety of
published detectors, considering
both several image databases, several attack strategies, and a few different experimental scenarios.

\subsection{Data Sets}
We experimented on the MNIST \cite{mnist} and CIFAR-10 \cite{cifar10} data sets.
MNIST is a ten-class dataset with 60,000 grayscale images, representing the digits `0' through `9'. 
CIFAR-10 is a ten-class dataset with 60,000 color images, consisting of various animal and vehicle categories.
Both data sets consist of 50,000 training images and 10,000 test images, with all classes equally represented
in both the training and test sets.  For AD purposes, the data batch under consideration in
our experiments consists of the test images plus the crafted attack images (whose generation is discussed
below).

\subsection{Classifiers} 
For training deep neural networks, we used mini-batch gradient descent with a cross entropy loss function
and a mini-batch size of 256 samples.
For MNIST, we trained the LeNet-5 convolutional neural-net \cite{LeCun}. This neural net reaches an accuracy of 98.1\% on the MNIST test set.
For CIFAR-10, we used the 16-layer deep neural network architecture proposed in \cite{He} and also used in \cite{CW}. This neural net, once
trained, reaches an accuracy of 89.47\% on the CIFAR-10 test set.

\subsection{Attacks}
During the phase of crafting adversarial samples, we only perturbed test set samples that were {\it correctly} classified. 
This is plausible since the attacker (who has full knowledge of the classifier, and thus whether a
sample is correctly or incorrectly classified by the classifier) is not likely to attack a sample that
the classifier is already misclassifying.
We implemented the fast gradient step method (FGSM) attack \cite{Goodfellow} and the Jacobian-based Saliency Map Attack (JSMA) \cite{Papernot} on the image data sets.  We also evaluated the CW attack, using the authors'
supplied code \cite{CW}.  
FGSM and CW are ``global'' methods, making small magnitude perturbations, but to all pixels in the image.  By contrast,
JSMA is a more ``localized'' attack, making changes to far fewer pixels, but with large changes needed on these
pixels, in order to
induce misclassifications.
For CW, we used the version that quantifies image distortion using the L2 metric, since we found this was the most challenging
CW version against which to defend.
For JSMA, we implemented the version that alters a (minimal set of) dark pixels, making them white. 
Moreover, while JSMA was only developed in \cite{Papernot} for gray scale images and only applied to MNIST there,
we extended their method to attack color images (CIFAR-10) by allowing perturbations of individual color planes, for each pixel,
with a maximum on the total number of such perturbations.  Again, perturbation of a pixel's color plane involved saturating
(to the maximum value).

For each test sample, from a particular class ({\it e.g.}, $c$), we randomly selected one of the other classes ({\it e.g.} $c'$), in an 
equally likely fashion, and generated an attack instance starting from the test image, using the given
attack algorithm, such that the classifier will assign the perturbed
image to class $c'$.  
In this way, for MNIST, we successfully crafted 9845 adversarial images using the JSMA attack (maximum percentage of pixels allowed to be perturbed is set to 13.75\%), 9762 adversarial samples using the FGSM attack(step size for gradient descent of 0.0005), and 9894 adversarial images using the CW-$L_2$ attack(Lagrange multiplier hyper-parameter set to 3).
For CIFAR-10, for the FGSM attack,
we successfully crafted 9243 adversarial images (step size for gradient descent of 0.001);
for the JSMA attack, we successfully crafted 9491 adversarial images (maximum percentage of pixel color planes allowed to be perturbed of 17.24\%); for the CW attack, we successfully crafted 9920 images (Lagrange multiplier set to 4).

\begin{figure}[h]\centering
\includegraphics[width=5in]{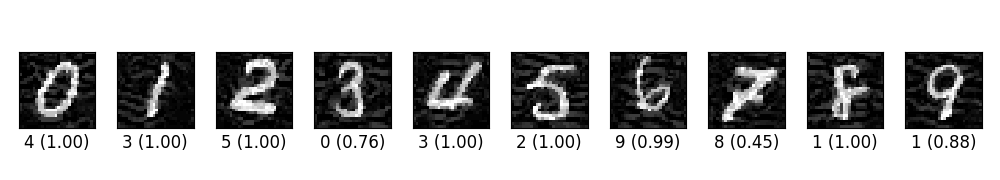}
\caption{CWL2 attack adversarial images.  Below each image the digit is the classifier's decision and the number in parentheses is the posterior probability under this winning class, indicating the decision confidence.}\label{fig:CW-array}
\end{figure}

Some attack examples are shown for FGSM, JSMA, and CW attacks on MNIST in 
Figures \ref{fig:FGSM-matrix}-\ref{fig:CW-array}.
Note that while FGSM is generally thought of as an ``imperceptible'' attack, there are `ghost' artifacts in
Figure \ref{fig:FGSM-matrix}
that are visually perceptible.  Likewise, for CW, there are ghost artifacts, especially horizontally oriented,
that are noticeable
(Figure \ref{fig:CW-array}).  
For CIFAR-10, though, we did indeed find FGSM and CW attacks to be 
visually imperceptible.  JSMA attacks, on the other hand, are quite visually perceptible in 
Figure \ref{fig:JSMA-matrix}
-- there
are extra white pixels and also visible salt and pepper noise.  Thus, while JSMA does induce misclassifications,
it is arguable whether the attack is fully ``successful'' in the sense defined in \cite{Papernot} in that the resulting images do have significant
artifacts that might cause a human being to misclassify the attacked image in some cases (or to profess ``I don't know'').
The latter is especially true for some images starting from the `5' and `3' categories\footnote{While \cite{Papernot} used a human subject study to confirm attack successes, they did not ask respondants
whether or not they thought the images had been tampered with.}.

We will also use and generalize the "security evaluation curve" concept from \cite{Biggio_wild} to show how effectively our detector works under different 
attack strengths.
For each attack, different parameters are varied to control the attack strength.
For JSMA, we vary the maximum number of pixels allowed to be modified for an image; for FGSM, we vary the step size used in gradient descent (a larger step may allow the attack to move further beyond the classifier's decision boundary); for CW, we 
vary the Lagrange multiplier that controls the L2 difference between the original and attacked images.
Security evaluation curves were proposed in \cite{Biggio_wild} to assess robust classification approaches, measuring 
classification accuracy versus attack strength.  For our AD approach, we appropriately extend this concept to assess 
a detection defense -- specifically, we evaluate the success rate in crafting misclassified images, detection performance,
and classification accuracy at a fixed true positive detection rate, all as a function of attack strength. 

Finally, we will also include a ``white box'' attack experiment, where the attacker has full knowledge of {\it both}
the classifier and the detector.
\subsection{Baseline Detectors}
We implemented the unsupervised null density detector from \cite{AD} and the blurring ``correction'' approach
from \cite{Li_ICCV}.  We used the code provided by the authors for the Openmax detector \cite{Openmax}.  We also
compare with results reported for the ten detectors evaluated in \cite{Wagner17}
(especially against the CW attack).

\subsection{Noisy Data Scenario}
The MNIST data is fairly ``clean'' -- the images are gray scale, but many pixels are nearly white or nearly black.
In order to model the scenario where the data is messier,
involving both evasive and {\it natural} adversaries (and thus where attack detection may be more challenging),
we also performed experiments where Gaussian noise was added to image intensity values.  For the case of FGSM attacks,noisy images were obtained by adding Gaussian noise to every pixel, with the mean and variance chosen to match
the mean and variance of the perturbations produced by the FGSM attack.
For the case of JSMA attacks, the same approach was applied (using the mean and variance estimated for the JSMA attack), but on a randomly chosen subset of pixels, whose
size was chosen to equal the number of pixels modified by the JSMA attack.  In this way, for both attacks, (noisy) non-attack images are 
generated that are more difficult to distinguish from attack images than the orginal (clean) non-attack images.

For experiments involving noisy images, the experimental protocol was as follows.  First, we designed classifiers and crafted attack
images for the clean data set.  Next, after estimating the mean and variance of the attack 
perturbations, we added noise to the training set.  We then retrained the classifier and estimated the class-conditional null densities based on the noisy training set.  We then crafted {\it new} attack images working from
the original test images -- this is required because the attack images should be successful in causing 
misclassification on the {\it new} classifier (trained on the noisy training images).  Finally, noise is
added to the original test samples, creating noisy (non-attacked) test samples\footnote{For MNIST, in the noisy case, the noisy classifier's test set accuracy was $0.972$ in the case of the attack from
\cite{Goodfellow} and $0.975$ in the case of the attack from \cite{Papernot}.  Thus, adding noise did not
substantially compromise the accuracy of classification.}.  The noisy test samples
and the new attack samples form the batch on which anomaly detection is performed. 
In the case of CIFAR-10, since the color images are intrinsically ``noisy'', we do not add any noise -- all CIFAR-10 experiments were thus based on the original training and test images.

\begin{figure}[h]\centering
\begin{tabular}{cc}
\begin{minipage}[t]{0.5\linewidth}
\centering\includegraphics[width=3in]{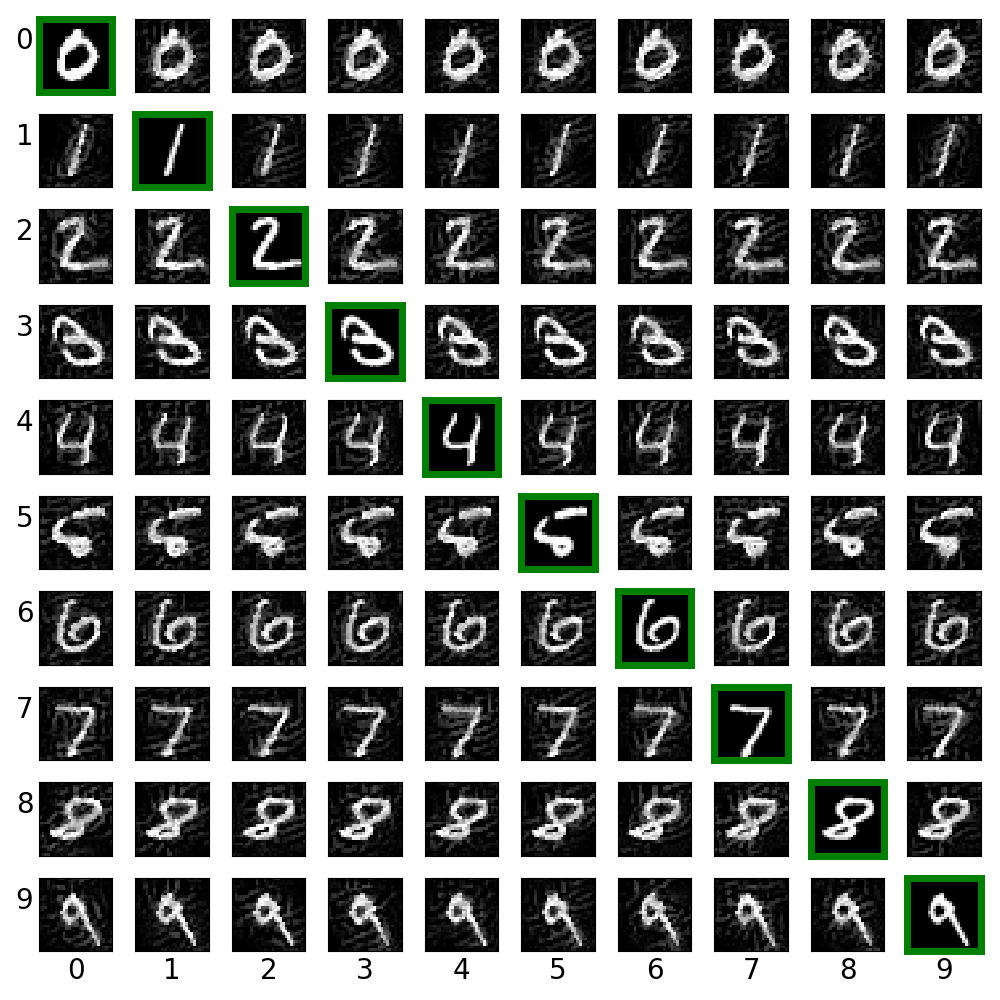}
\caption{FGSM adversarial image matrix, with starting true class on the row and misclassified class on the column.  The main diagonal shows non-attacked, correctly classified images.}\label{fig:FGSM-matrix}
\end{minipage}
\begin{minipage}[t]{0.5\linewidth}
\centering\includegraphics[width=3in]{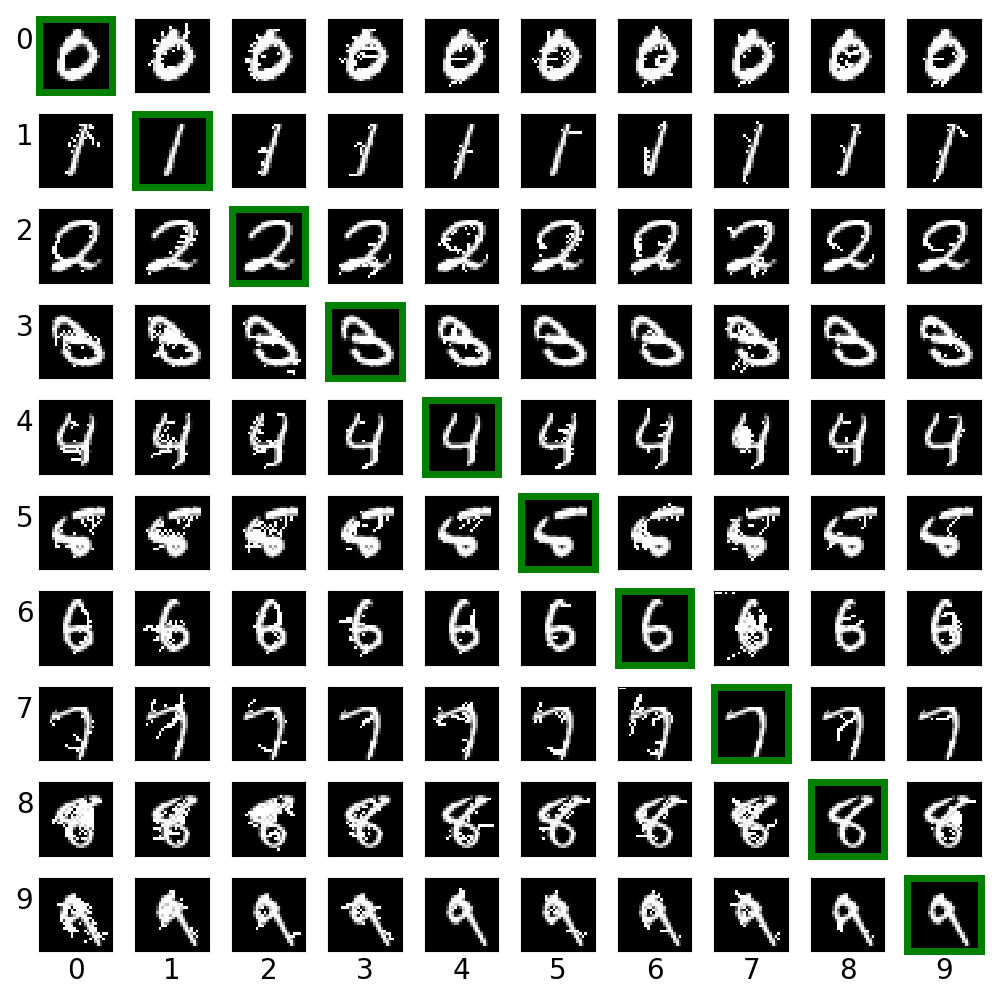}
\caption{JSMA adversarial image matrix.}\label{fig:JSMA-matrix}
\end{minipage}
\end{tabular}
\end{figure}

\subsection{Null Density Modelling}
For modelling null densities for several deep layers, we considered both the Gaussian kernel density estimator used in \cite{AD},
Gaussian mixture models (GMMs), and multivariate log-normal mixtures as described in the
previous section.   The number of mixture components was chosen for GMMs
and the log-normal mixtures to minimize the Bayesian
Information Criterion.  For the Gaussian kernel density, the variance parameter was chosen to maximize
likelihood on the training set.  For the GMM and log-normal mixtures, we considered both full covariance matrices and diagonal
covariance matrices, depending on the dimensionality of the DNN layer being modelled.
Table \ref{table:lenet5_or_16}
below shows the layer-dependent GMM modelling choices we made.
Note that Lenet-5 has one max-pooling layer while the 16-layer neural-net has 2 max-pooling layers. 
\begin{table}[h]
\centering
\begin{tabular}{|c|c|c|} \hline
 & penultimate layer & max-pooling layer \\ \hline
 Lenet-5 for MNIST & full covariance & diagonal covariance \\ \hline
 16-layer DNN for CIFAR-10 & full covariance & diagonal covariance \\ \hline 
\end{tabular}
\caption{GMM modeling}\label{table:lenet5_or_16}
\end{table}

\subsection{Anomaly Detection Scenarios}
In our experiments, there are essentially 3 different experimental scenarios that were investigated in
the ``black box'' case, where the attacker does not have knowledge of the detector: 
\begin{itemize}
\item {\it clean case}: We do not craft any noisy samples. The training phase is based on the original training set and the test batch for AD consists of the original test samples and the crafted adversarial samples. All the experiments for CIFAR-10 were done under this case.
\item {\it noisy case}: The experimental protocol for this case was previously discussed.   The test batch for
AD in this case consists of noisy versions of the original test samples (essentially, natural
adversaries) and crafted adversarial samples that
induce misclassifications by the classifier that was trained on the noisy training set.  Note that the detection
problem is expected to be more difficult here, compared with the clean case, as will be borne out by our results. 
\item {\it mismatch case}: noisy is only added to the original test samples, while training is still based on the original (clean) training set. In this case, 
we included in the AD test batch the original (clean) test images, the noisy test images, and the adversarial
images.
Because the classifier and null modeling are based on clean data whereas noisy data is included in the AD test
batch, we refer to this as the ``mismatched'' case.  This case is useful for assessing system robustness
(when training and test conditions are statistically mismatched).
\end{itemize}
\subsection{Experiments Involving Several ADA Variants}

\begin{figure}[h]\centering
\begin{tabular}{cc}
\begin{minipage}[t]{0.5\linewidth}
\centering\includegraphics[width=3in]{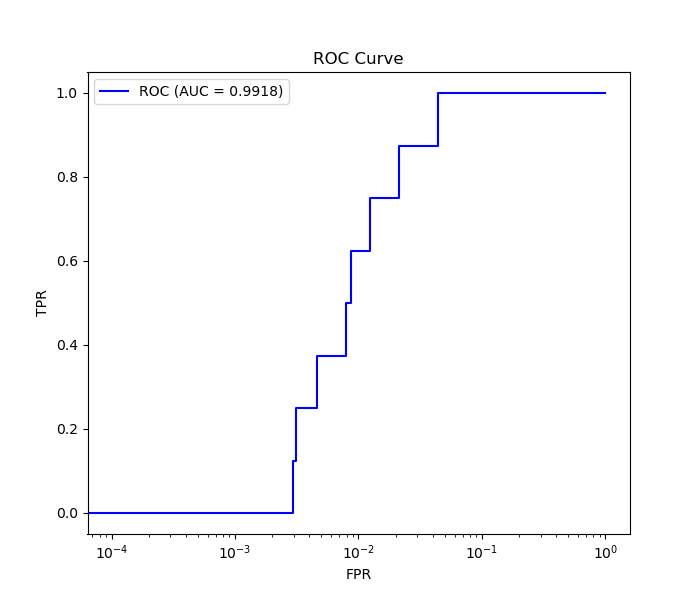}
\caption{ADA-GMM ROC curve, MNIST, FGSM attack, clean case.}\label{fig:ADA-GMM-MNIST-FGSM-clean}
\end{minipage}
\begin{minipage}[t]{0.5\linewidth}
\centering\includegraphics[width=3in]{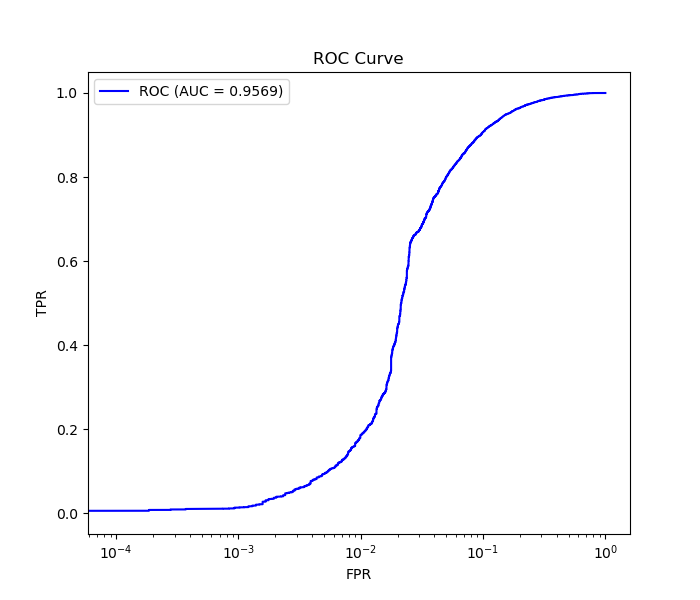}
\caption{ADA-maxKL ROC curve, MNIST, FGSM attack, noisy case.}\label{fig:ADA-maxKL-MNIST-FGSM-noisy}
\end{minipage}
\end{tabular}
\end{figure}

\begin{table}[h]
\centering
\begin{tabular}{|c|c|c|c|c|c|} \hline
  & ADA-kernel & kernel in \cite{AD} & ADA-GMM & ADA-maxKL & AWA-ADA-maxKL \\ \hline
 clean & 0.9703 & 0.9746 & 0.9918 & NA & 0.9892\\ \hline
 noisy & 0.9049 & 0.8752 & 0.8695 & 0.9569 & 0.9554 \\ \hline
 mismatch & 0.9201 & 0.7825 & NA & NA & NA \\ \hline
\end{tabular}
\caption{ROC AUC score on MNIST dataset with FGSM attack}\label{table:MNIST_under_FGSM}
\end{table}
Table \ref{table:MNIST_under_FGSM}
shows results on MNIST under the FGSM attack. In the first 3 columns, we only null-modeled the penultimate layer of the DNN.  For ADA-maxKL (applied in the noisy case), maxKL is based on two layers -- the single maxpooling layer and the penultimate layer.  The penultimate layer is modeled using a GMM with a full covariance matrix while the maxpooling layer is modeled by a Gaussian kernel.  
Note that all the methods work very well for the clean MNIST data set, but with ADA-GMM giving the best results
and a highly compelling 0.992 AUC (including a point with (TPR,FPR) = (0.94, 0.03)).  
For the noisy case, the maxKL paradigm, considering anomalies in two different
layers, is needed to get the best results for ADA.  This method significantly improves over the other ADA methods
(excepting AWA-ADA-maxKL)
and over \cite{AD}, achieving 0.957 AUC (and a point with (TPR,FPR) = (0.91, 0.06)). For these two (best) results, we show the ROC curves in 
Figures \ref{fig:ADA-GMM-MNIST-FGSM-clean} and \ref{fig:ADA-maxKL-MNIST-FGSM-noisy}.
 For the mismatch case, our basic ADA-kernel method, which exploits null information from both $c_d$
and $c_s$, is seen to substantially outperform \cite{AD}, which only exploits information 
from $c_d$.

We also note that since we propose to detect and then classify if there are no detections, our system 
changes the distribution of the non-attack samples being classified (only those not falsely detected
as attacks), which could affect accuracy of the classifier.   However, we have found that at relatively
modest FPRs (e.g. 5\% or less), there are extremely modest changes in the classifier's
(conditional) test set accuracy (based on the test set that excludes false detections). This is true for both MNIST and CIFAR-10.  Related results are reported later.

Note that, beyond exploiting $c_s$ and $c_d$, it is possible to define probability vectors on the {\it full}
complement of classes, with KL distance measured between these probability vectors. 
Table \ref{table:CIFAR-FGSM-ADA-kernel}
shows the AUC difference between just using 2 classes (source and destination) and using all classes. In this case, we applied the ADA-kernel detector in penultimate layer modelling on the CIFAR-10 dataset.  In this experiment and, in general, anecdotally, we have found that the gains in going from use of $c_s$ and $c_d$ to use
of {\it all} classes is typically modest, with the gains in going from use of just $c_d$ \cite{AD} to use of
both $c_s$ and $c_d$ often much greater.  This validates 
the main idea of the ADA detection paradigm -- that an attack example is expected to be ``too atypical'' with respect to
$c_d$ and ``too typical'' with respect to $c_s$; thus, good AD performance should be achievable by exploiting 
anomaly signatures measured with respect to both $c_s$ and $c_d$, rather than just $c_d$ \cite{AD}.
\begin{table}[h]
\centering
\begin{tabular}{|c|c|c|} \hline
  & two classes & all classes   \\ \hline
 clean & 0.8159 & 0.8273 \\ \hline
\end{tabular}
\caption{AUC score on CIFAR-10 dataset with FGSM attack and ADA-kernel method.}\label{table:CIFAR-FGSM-ADA-kernel}
\end{table}

\begin{table}[h]
\centering
\begin{tabular}{|c|c|c|c|c|} \hline
  & AW-ADA-maxKL & ADA-maxKL & ADA-kernel & kernel in \cite{AD}  \\ \hline
 clean & 0.9235 & 0.8756 & 0.8289 & 0.8273  \\ \hline
 ideal & NA & 0.9155 & NA & NA \\ \hline
\end{tabular}
\caption{AUC score on CIFAR-10 dataset with FGSM attack.}\label{table:CIFAR-FGSM}
\end{table}
Table \ref{table:CIFAR-FGSM}
shows results on the CIFAR-10 dataset under the FGSM attack.  
AUCs are much lower than for MNIST.  One possible reason is that the classes are much more 
confusable for CIFAR-10 (with only 0.89 test set accuracy) than for MNIST.  However, the maxKL paradigm
still gives substantial AUC {\it gains} over both \cite{AD} and the basic, single layer ADA method.
Moreover, the use of confusion matrix and source uncertainty information by AW-ADA-maxKL gives significant further
improvement, from 0.8756 up to 0.9235 AUC (including a point with (TPR,FPR) = (0.8, 0.1)). 

The ``ideal'' row shows results for an experiment where the {\it misclassified} test samples are excluded
from the AD test batch.  While these samples cannot of course be excluded in practice, the goal here is to understand whether or not {\it most} of the ADA-maxKL suboptimality
on CIFAR-10
is attributable to misclassified test samples.  While the AUC does increase under the ``ideal'' case,
it is still well below 1.0 (and below AW-ADA-maxKL's AUC in the clean case).

For ADA-maxKL in the clean case in 
Table \ref{table:CIFAR-FGSM},
we modeled the penultimate layer using a GMM and modeled the remaining two maxpooling layers with a Gaussian kernel.
\begin{table}[h]
\centering
\begin{tabular}{|c|c|c|} \hline
  & K-K-G & K-K-K  \\ \hline
 clean & 0.8756 & 0.8567 \\ \hline
\end{tabular}
\caption{AUC score on CIFAR-10 dataset with FGSM attack and ADA-maxKL.}\label{table:CIFAR-FGSM-ADA-maxKL}
\end{table}
We also want to illustrate how different modelling choices for these layers affect performance. For concision, we use G as short for GMM and K as short for Gaussian kernel, i.e., K-K-G means the first 2 maxpooling layers are modeled using a Gaussian kernel while the penultimate layer is modeled using a GMM.  
Table \ref{table:CIFAR-FGSM-ADA-maxKL}
indicates that how
the penultimate layer is modeled has some albeit a modest effect on the detection accuracy on CIFAR-10.
\subsection{Experiments Involving the JSMA Attack on MNIST}
\begin{table}[h]
\centering
\begin{tabular}{|c|c|c|c|c|c|c|} \hline
  & AW-ADA-maxKL & ADA-maxKL & ADA-kernel & L-AWA & white-counting & region counting \\ \hline
 clean & 0.9397 & 0.9314 & NA & 0.9893 & 0.9466 & 0.97 \\ \hline
 noisy & 0.8961 & 0.8908 & 0.8807 & 0.9325& 0.9062 & NA \\ \hline
\end{tabular}
\caption{AUC scores of various anomaly detectors on MNIST under the JSMA attack.}\label{table:various-detectors-MNIST-JSMA}
\end{table}

Table \ref{table:various-detectors-MNIST-JSMA}
evaluates several methods on the JSMA attack applied to MNIST in the clean and noisy
cases, with the maximum fraction of modified pixels set to $9\%$.  
Note that ADA models the {\it joint} feature vector for a layer, which is a function of the
{\it entire} image.  Thus, we would expect that ADA is most suitable for detecting {\it global} attacks,
wherein many/most pixels in the image are modified -- this is borne out by our very strong detection
results on MNIST for FGSM attacks.  JSMA, however, strongly restricts the number of modified pixels
(but necessitates gross changes be made to these pixels, in order to succeed in inducing miclassifications).
JSMA, accordingly, is a more ``local'' attack.  Thus, we might expect ADA not to perform as well in detecting
JSMA attacks as FGSM attacks.  This is borne out in 
Table \ref{table:various-detectors-MNIST-JSMA},
where AWA-ADA-maxKL manages a respectable 0.9397
AUC in the clean case (but not nearly the 0.992 AUC achieved in detecting FGSM attacks).  However, this is not
to say that JSMA attacks are intrinsically harder to detect.  In fact, looking at the images in Figure \ref{fig:JSMA-matrix},
the JSMA attacks are quite discernible, with salt and pepper noise and extra white content, which should
make them readily detectable (by a simple, but suitably defined detector).  To demonstrate this, we constructed
two very simple detectors for JSMA attacks.  One forms the class-conditional (null) histogram on the number of white
pixels in the image.  To make a detection decision on a given image, one counts the number of white pixels,
$N_w$, identifies the class whose mean white count is closest to $N_w$, and then computes a one-sided p-value
at $N_w$, based on the class's (null) histogram.  This p-value is the decision statistic that is thresholded.
This very simple detector achieves an AUC of $0.946$ in the clean MNIST case, a bit better than ADA-maxKL.
To achieve even better results (specifically targeting the clean image case), one can simply count the
number of disjoint white regions in the image.  Clean MNIST digits typically consist of a {\it single}
white region (where a region is defined as a collection of pixels that are ``connected'', with two white
pixels connected if they are in the same first-order (8-pixel) neightborhood, and with the region defined
by applying transitive closure over the whole image).
By contrast, nearly all JSMA attack images have extra
isolated white regions (associated with salt and pepper noise).  Simply using the number of white regions
in the image as a decision statistic yields 0.97 AUC in the clean MNIST case -- this strong result
for this very simple detector indicates the susceptibility of the JSMA attack to a (simple) AD 
strategy, even as the ADA approaches evaluated until now are not most suitable for this attack.
\subsection{Comprehensive Results for L-AWA-ADA-maxKL}
Two experimental results given so far motivate the need for our {\it best} detection approach, L-AWA-ADA-maxKL:
1) AWA-ADA-maxKL
achieves only 0.94 and 0.896 AUCs in the clean and noisy cases, respectively, for JSMA on MNIST;
2) Although we have not yet discussed CW attack results, against the CW attack on CIFAR-10, AWA-ADA-maxKL
achieves 85\% TPR
at 28\% FPR and 81\% TPR at 23\% FPR.  
While this is better than the best (TPR,FPR) pair results reported for ten detection methods in \cite{Wagner17},
L-AWA-ADA-maxKL will be seen to 
substantially improve on these results.  
Thus, in this section, we give a comprehensive performance evaluation for L-AWA-ADA-maxKL, including:
1) performance as a function of attack strength 
(``security evaluation curves'') 
for L-AWA;
2) comparison of L-AWA against several baseline detectors;
3) comparison with the detection results reported in \cite{Wagner17} for the CW attack;
4) results for a white box attack on L-AWA. 

First, as shown in 
Table \ref{table:various-detectors-MNIST-JSMA},
L-AWA-ADA-maxKL achieves a 
quite compelling AUC of 0.9893 in the clean case and 0.9325 in the noisy case, for JSMA on MNIST.   Thus, L-AWA-ADA-maxKL substantially outperforms the AWA-ADA-maxKL results in 
Table \ref{table:various-detectors-MNIST-JSMA}
(and even outperforms
the region-counting detector). 
L-AWA in the clean case achieves a point with (TPR,FPR) = (0.91, 0.07).
Moreover, for the FGSM attack on CIFAR-10, L-AWA achieves an AUC of 0.9265, slightly better than that reported for
AWA-ADA-maxKL in 
Table \ref{table:CIFAR-FGSM}.

\begin{figure}[h]\centering
\centering\includegraphics[width=3.5in]{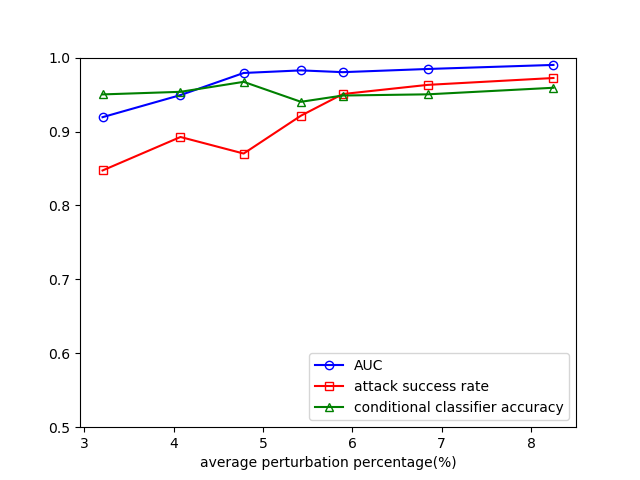}
\caption{L-AWA-ADA-maxKL detection results for the JSMA attack on 
MNIST.}\label{fig:L-AWA-ADA-maxKL-JSMA-MNIST}
\end{figure}
	
\paragraph{Security Evaluation Curves for JSMA on MNIST:} 
In 
Figure \ref{fig:L-AWA-ADA-maxKL-JSMA-MNIST},
on the y-axis, we plot L-AWA's ROC AUC, the attack success rate, and the conditional classifier accuracy (evaluated
only on the non-attack images which are not falsely detected, for a TPR of 0.8125 -- we fixed the TPR as the attack
strength is varied, while the FPR varies with attack strength).  These are plotted as a function 
of the
average fraction of pixels modified on each image by JSMA (which is a function of the attack strength -- the {\it maximum} allowed fraction for an image). 
Note the attack success rate tends to increase with attack strength, as expected.  But this makes the attack highly detectable --
AUCs above 0.98 are achieved when the average modified fraction is 5\% or above.
Note also that the (conditional) classifier accuracy (which varies because the FPR varies with the attack strength)
exhibits quite small variations over the
range of explored attack strengths.   

\begin{figure}[h]\centering
\centering\includegraphics[width=3.5in]{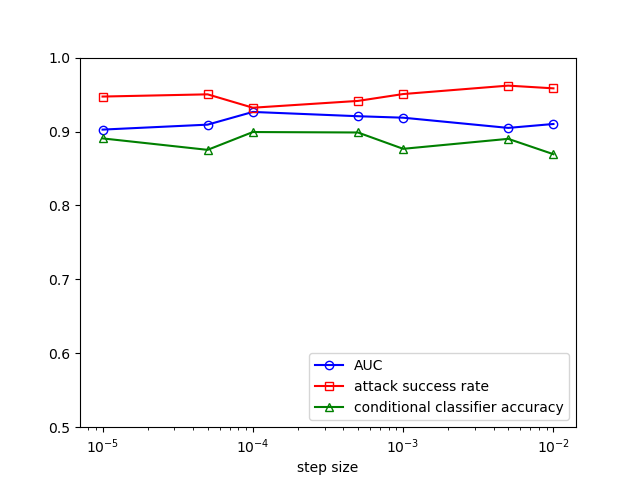}
\caption{L-AWA-ADA-maxKL detection results for the FGSM attack on 
CIFAR-10.}\label{fig:L-AWA-ADA-maxKL-FGSM-CIFAR}
\end{figure}

\paragraph{Security Evaluation Curves for FGSM on CIFAR:}
Figure \ref{fig:L-AWA-ADA-maxKL-FGSM-CIFAR}
shows 
the same types of curves as in 
Figure 
Figure \ref{fig:L-AWA-ADA-maxKL-JSMA-MNIST}
for the FGSM attack on CIFAR-10, with the FGSM gradient's step size playing
the role of attack strength.
We see that none of the three curves shows a clear trend as the FGSM step size is varied.  This is not so surprising,
since the step size may only weakly control the attack ``strength'' -- as the step size increases, there may be {\it some} 
tendency to produce attacked images that go further past the classifier decision boundary than when the step size is 
smaller.  However, this effect may be quite modest, which explains why the ROC AUC and attack success rate remain
roughly flat as the step size is varied. 
We also report that, at an FPR of 0.0845, 15.92\% of misclassified images are (falsely) detected as attacks.  Thus, the false detection rate is higher in the population of misclassified images than in the overall
image population.  This is plausible, as misclassified images may ``look'' more like attacked images than
correctly classified images.

\begin{figure}[h]\centering
\centering\includegraphics[width=3.5in]{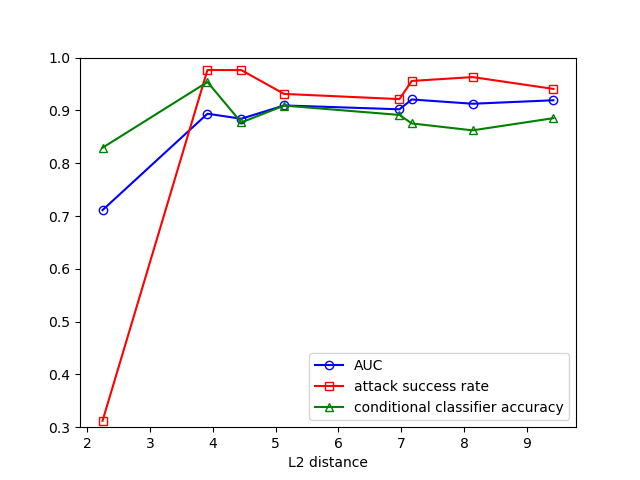}
\caption{L-AWA-ADA-maxKL detection results for the CW L2 attack on 
CIFAR-10.}\label{fig:L-AWA-ADA-maxKL-CWL2-CIFAR}
\end{figure}

\paragraph{Security Evaluation Curves for CW on CIFAR:}
The third attack investigated is CW \cite{CW} 
(Figure \ref{fig:L-AWA-ADA-maxKL-CWL2-CIFAR}).
Here, the attack strength is controlled by a Lagrange 
multiplier that determines the average L2 distance between the original and
attacked images.
In this case, we chose 8 Lagrange multiplier values, $\{0.5,1,2,5,10,15, 25,50\}$, with the L2 image
distortion increasing with this attack strength parameter. 
For CW, just as for JSMA, there are clear trends as the attack strength is varied.  In particular,
both the attack success rate and the detection accuracy tend to increase with the Lagrange multiplier --
for a Lagrange multiplier value of 0.5,
the detection performance is poor (AUC=0.7112), but the attack success rate is also quite low (0.3121). 
We also report that, at an FPR of 0.1216, 19.27\% of misclassified images are falsely detected as attack --
again misclassified images are over-represented in the population of false positive images.

\paragraph{Comparison With Other Detection Approaches:}
Here we compare L-AWA with two other detectors, given a fixed attack strength: 
the ``correction-based'' detector using image blurring from \cite{Li_ICCV} and the Openmax detector \cite{Openmax}.  We
also compare with results from \cite{Wagner17}.
As mentioned earlier, Openmax requires setting two hyperparameters using labeled attack examples.  Thus, this
method is actually supervised.
\begin{figure}[h]\centering
\begin{tabular}{cc}
\begin{minipage}[t]{0.5\linewidth}
\centering\includegraphics[width=3in]{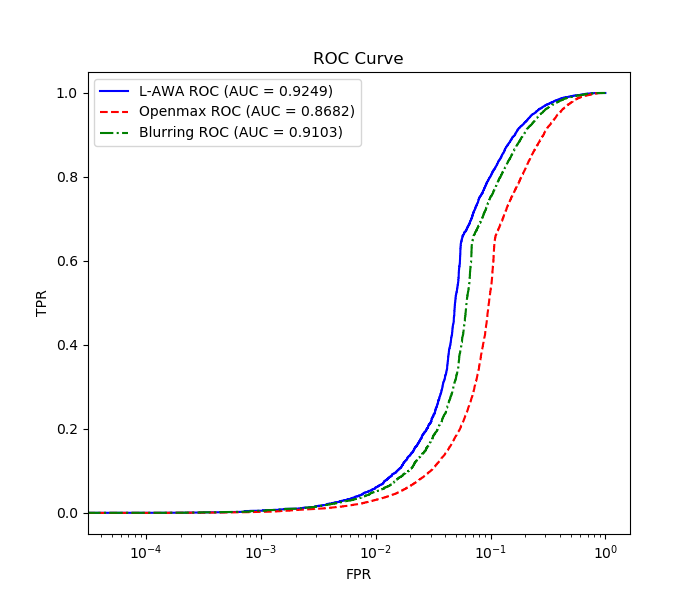}
\end{minipage}
\begin{minipage}[t]{0.5\linewidth}
\centering\includegraphics[width=3in]{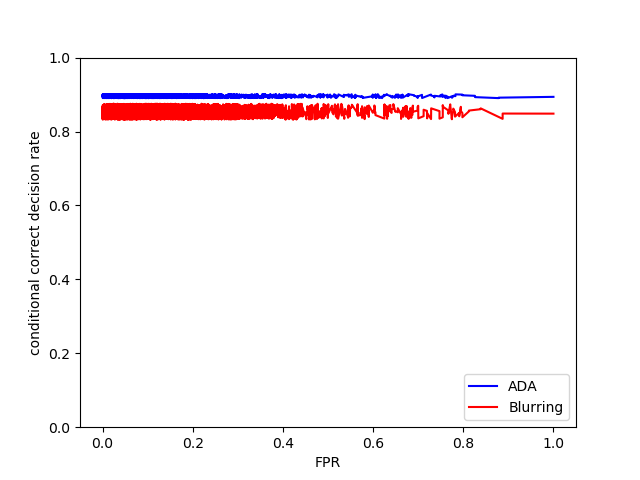}
\end{minipage}
\end{tabular}
\caption{Detector Comparison for the FGSM attack on CIFAR-10.}\label{fig:compare-FGSM-CIFAR}
\end{figure}
Figure \ref{fig:compare-FGSM-CIFAR}
shows the performance of the three detectors against the FGSM attack on CIFAR-10. In this case, we used 137 adversarial images to determine the hyper-parameters for the Openmax method. We can see that although the blurring method is quite competitive in its detection performance with L-AWA, this is achieved with some compromise in the (conditional) classification accuracy as the attack FPR is varied. 
Openmax's detection accuracy is the worst, despite its use of (quite a few) labeled attack examples
for setting its hyperparameters.

We also assessed the execution time required by methods to make detections.
Currently we are using two NIVIDA GTX1080 GPUs, an Intel Core i7-5930K Processor and a 256GB DDR4. For this platform, detection time for a CIFAR-10 image using L-AWA is 0.00114s and 0.00063s for the blurring method, {\it i.e.} it is both quite modest and
comparable for the two methods.  For comparison's sake, the average time required by the attacker to craft a successful CW attack example
on CIFAR-10 (measured over the whole test set) is 0.72s -- $\sim$ 700 times that required for L-AWA detection.
Thus, detector computational complexity is much lower than that of the attacker.

\begin{figure}[h]\centering
\begin{tabular}{cc}
\begin{minipage}[t]{0.5\linewidth}
\centering\includegraphics[width=3in]{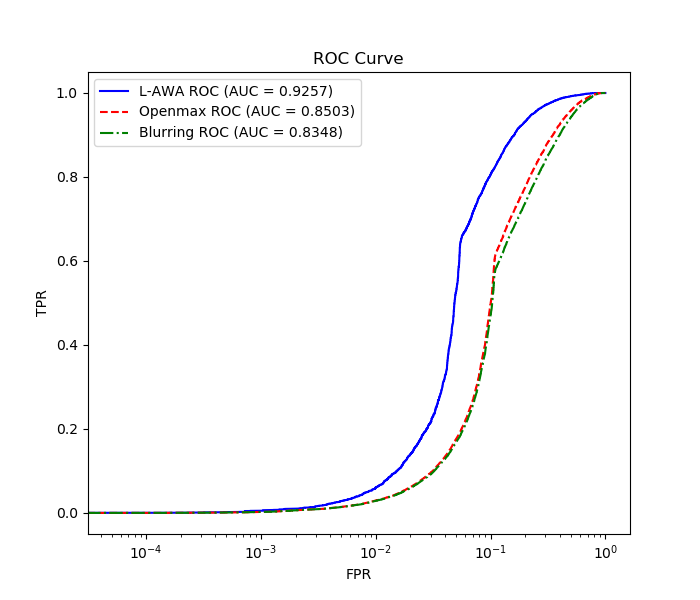}
\end{minipage}
\begin{minipage}[t]{0.5\linewidth}
\centering\includegraphics[width=3in]{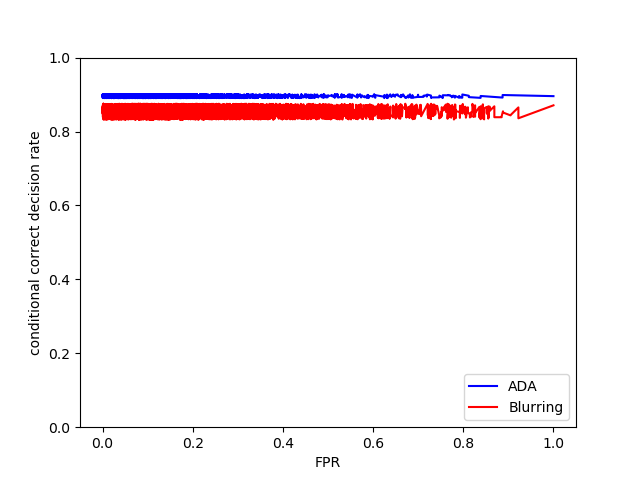}
\end{minipage}
\end{tabular}
\caption{Detector Comparison for the CW L2 attack on CIFAR-10.}\label{fig:compare-CWL2-CIFAR}
\end{figure}
\begin{figure}[h]
\centering
\includegraphics[width=4in]{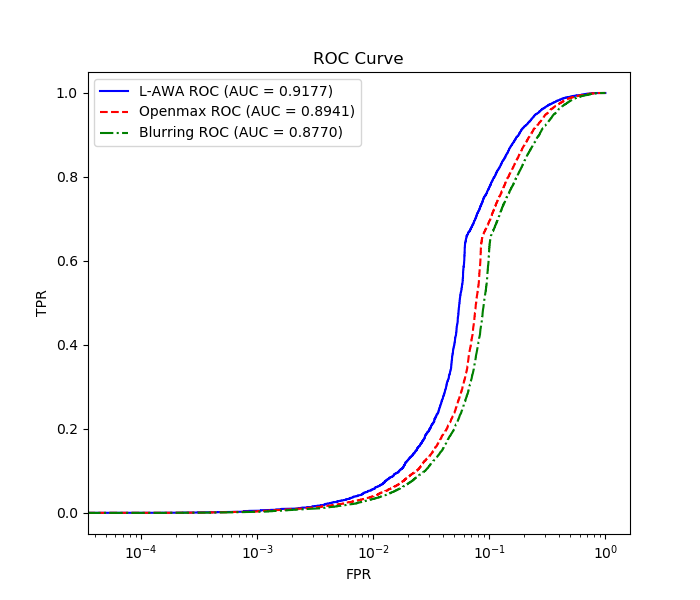}
\caption{Detector comparison for the JSMA attack on CIFAR-10.}\label{fig:compare-JSMA-CIFAR}
\end{figure}

For the CW L2 attack 
(Figure \ref{fig:compare-CWL2-CIFAR}),
we used 152 adversarial samples to determine the  hyper-parameters for Openmax. 
L-AWA clearly outperforms both baseline detectors in this experiment.
Again, the blurring method compromises (conditional) classification accuracy (and has more variability
in this accuracy), as the FPR is varied.
Finally, for the JSMA attack on CIFAR-10 
(Figure \ref{fig:compare-JSMA-CIFAR}), 
with maximum number of modified pixel planes of
9.7\%, we again see that L-AWA outperforms the two baseline methods, even though the JSMA attack on CIFAR-10 (unlike MNIST) is not so easy to detect for any of the methods.  In this case, the Openmax method used 250 labeled attack images to set its hyperparameters.
Also, in this case, the blurring method performs the worst. 

\paragraph{Results for the CW attack on CIFAR-10:}
As noted earlier,
the best approach reported in \cite{Wagner17} against CW was a {\it supervised} detection approach \cite{Metzen}, achieving
81\% TPR at a 28\% FPR\footnote{\cite{Wagner17} does report higher detection rates for a few other methods.  However,
in these cases, FPR was not reported.}.  The unsupervised AW-ADA-maxKL detector (using GMM densities) achieves 85\% TPR at 28\% FPR
and 81\% TPR at 23\% FPR.
L-AWA (using multivariate log-normal mixture densities) achieves what we believe is a
current state-of-the-art result against CW:
0.9107 AUC, 81\% TPR at 12\% FPR, and 96\% TPR at 28\% FPR -- better than results reported in \cite{Wagner17} and also significantly better than (global) AW-ADA-maxKL.  There of course is still potentially room
for further improvement, as this is far from perfect detection performance.

\paragraph{Deep Layer Atrribution:}
We also mention that, for L-AWA-ADA-maxKL on CIFAR-10, all 3 modelled deep layers contributed significantly to the decisionmaking:
the first maxpooling layer was the winner 37\% of the time, the second maxpooling layer won 40\% of the time, and
the penultimate layer won 23\% of the time (averaged over all detections, for all three attacks).

\subsection{L-AWA-ADA-maxKL Against a Fully White Box Attack}

\begin{figure}[htb]
\includegraphics[width=4in]{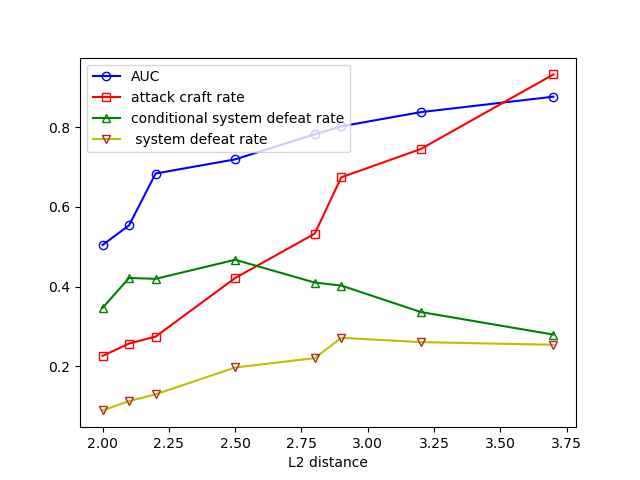}
\caption{CW white box attack results on CIFAR against L-AWA detection.} \label{fig:whitebox}
\end{figure}

We extend the white box attacker's objective function from \cite{Wagner17}, devising
an attack to defeat both the classifier and the detector.  Specifically, the attacker chooses the
image perturbation $\underline{\delta}$ seeking to solve: 
${\rm min}_{\underline{\delta}} \left(||\underline{\delta}||_2+c\cdot(f(\underline{x}+\underline{\delta})+D(\underline{x}+\underline{\delta}))\right)$. 
Here, 
$f(\cdot)$ and $D(\cdot)$ are loss functions as defined in \cite{Wagner17},
$f(\cdot)$ non-negative when a correct classification is made and $D(\cdot)$ non-negative when a detection is made,
$D(\cdot)$ using knowledge of the detector's threshold on its decision statistic.  This (additive)
cost is minimized when a modestly perturbed image is both misclassified {\it and} not detected.
The detector's FPR was set to 14\%.  The attacker cycles over the whole image, pixel by pixel, over all
three color planes, using a perturbation step of 0.05.  A perturbation is accepted if the cost function
decreases. The construction is stopped when cost function changes fall below $10^{-4}$ for two consecutive image passes.
In Figure \ref{fig:whitebox} we plot: the success rate in crafting misclassified images (the attack craft rate), the
detector's ROC AUC, the fraction of successfully crafted images that are not detected (the conditional system defeat
rate), and the fraction of attempted attacks that are both misclassified {\it and} not detected (the system defeat rate), all as a 
function of attack strength (L2 distortion).  As in previous black box experiments, the attack craft rate and the attack's detectability (ROC AUC) increase with attack strength.  Looking at the conditional system defeat rate,
there is an attack strength ``sweet spot'' of 2.5 where more than 40\% of crafted images are not detected.
However, {\it less than} 40\% of attempted attacks defeat the classifier at this attack strength.  The ultimate 
figure of merit, the system defeat rate, has a maximum value of $\sim$ 0.25 at an attack strength of $\sim$ 2.9 --
per white box attack attempt, the maximum attack success rate is about 25\%.  This is higher than
in the black box case (the CW attack success rate in defeating L-AWA was 1-0.81 = 0.19 at 12\% FPR),
but only modestly so.  Even with white box knowledge, it is not so easy to craft images to defeat
both the classifier and (L-AWA) detector.  Moreover, as noted earlier, these attacks are much heavier, computationally,
than the detection (or detection plus classification) effort. 

\section{Discussion and Future Work Directions}
We have only investigated the attacks from \cite{Goodfellow}, \cite{Papernot}, \cite{CW}, as well as a white box attack related to \cite{Wagner17}.  There is quite
interesting work on {\it universal} attacks, {\it i.e.} a {\it single} perturbation vector/image that induces
classification errors when added to {\it most} test patterns for a given domain \cite{Dezfooli}.  We believe this work is important because it is suggestive of the fragility
of DNN-based classification.  However, a single perturbation vector that is required to induce classification errors
on nearly all test patterns from a domain must be ``larger'' than that required to successfully misclassify
a single test pattern (and the approaches from \cite{Goodfellow},\cite{Papernot},\cite{CW} customize
the perturbation vector for each individual test pattern).  
Thus,
our approach,
which we have demonstrated to be quite successful in detecting the attacks from
\cite{Goodfellow} and \cite{Papernot}, should also be highly successful in detecting
universal attacks.

On the other hand, while our approach exploits deep layer attack signatures in DNNs, there are also recent
attacks that effectively seek to {\it destroy} such signatures (beyond approaches such as \cite{Li_ICCV}).
The attack in \cite{Sabour} was not
explicitly designed to induce (imperceptible) misclassifications -- rather to
make imperceptible changes to an image, $X$, such that the representation in layer $l$,
$\underline{g}_l(X + \Delta)$ will be very similar to the representation $\underline{g}_l(X')$ for 
a different image, $X'$.
If the images $X$ and $X'$ come from two different classes, the resulting perturbation may induce
the classifier to misclassify the perturbed image to the class of the target image.
In principle, this could form the basis for a good {\it grey box} attack,
simply exploiting knowledge of which DNN layers are used for AD and applying the attack to those
layers.
However, this attack was not
successful on either MNIST or CIFAR-10 \cite{Sabour}.  This is actually not so surprising
 -- it is very stringent to force
an internal (deep) layer for the perturbed image to agree closely with a different (target) image's 
representation of that layer
-- the two representations need {\it not} necessarily be very similar to  
induce a misclassification.  
That is, forcing such agreement may cause misclassifications, but the required perturbations may need to 
be much larger than those required for attacks with objectives more directly related to misclassification such
as \cite{Goodfellow},\cite{Papernot},\cite{CW}.
Accordingly, \cite{Sabour} could not
reliably make inobtrusive misclassifications on MNIST and CIFAR-10.  However, some variation on \cite{Sabour}, better matched
to the misclassification objective, could potentially form a challenging grey
box attack on our AD.  This could be investigated in future.

Another promising research direction concerns reverse engineering attacks.  Test-time 
evasion attacks rely on knowledge of the classifier and a reverse engineering attack may first
need to be deployed to obtain this knowledge.  We earlier noted in the Introduction that detection of
reverse engineering based on random queries \cite{Reiter} should be very easy.  However, more
recent reverse engineering attacks \cite{Papernot3} generate more realistic queries that are perturbations
of known examples from the domain.  It is possible AD techniques such as ours may
be effective at detecting (realistic) reverse engineering queries, just as they have
capability to detect test-time evasion attacks.  This may be investigated in future. 

While our approach is unsupervised, we could investigate a supervised variant
that uses our local ADA-AWA statistics as input features to a classifier, which is 
trained on some ``known''
attacks and tested on held-out ``unknown'' attacks.  This may also be investigated in future.

While we have focused on DNN classifiers, our approach is suitable more generally to complement other
classifiers in detecting test-time attacks.  
Likewise, we have only considered image classification domains.  Our approach should be suitable
for other domains such as speech recognition \cite{Wagner16} or music genre classification \cite{Larsen}.  
Here, the feature vector might not be the raw data (speech waveform) - it might be cepstral or wavelet coefficients.
Accordingly, the null density models may need to be well-customized for this domain, in order to give the strongest
attack detection results ({\it e.g.}, hidden Markov models may be needed here).
Moreover, our approach can also be readily applied when features are categorical, ordinal, or mixed continuous-discrete.
For discrete feature spaces ({\it e.g.}, a {\it text} domain, where the attacker is
seeking to (imperceptibly) modify a document to fool a document/email classifier),
(null) density functions would need to be replaced by (sufficiently rich) joint probability mass function
null models.
All of these are potential directions to consider in our future work.

\section{Conclusions}
We have strongly argued here for an (in general, unsupervised) AD approach to detect test-time
evasion attacks, to be used in conjunction with a classifier.  While there are applications
and scenarios where simply 
seeking to correctly classify in the face of the attack is appropriate,
we have identified attack scenarios where, in the presence of an attack,
correct classification has no utility (it is solely for the benefit of the attacker).
Moreover, in security-sensitive settings attack detection may help to drive attack 
mitigation protocols, as well as conservative actions (safely stopping a self-driving vehicle,
rescanning a medical image).
Even for classification domains where human perception is {\it not} a factor -- e.g.
huge-dimensional
feature space domains (gene microarray or other bioinformatics domains, large-scale sensor array (e.g.
Internet-of-things) domains, documents and video (unless a human will actually read the whole document or watch the whole video)) or domains whose
feature vectors are essentially ``voodoo'' (uninterpretable) to a human being (e.g., computer software, for malware
detection) -- one could argue that an attacker will try to minimize the attack ``footprint'' to evade detection,
and thus a robust classifier may still be able to correctly classify the given content, even in the
presence of the attack.  However,
it may be less important to classify the software type of a computer program if the program in fact
contains malware.  Likewise, if the goal of an attack was not to fool a classifier, but rather simply to embed
false content (into a document or video), then detecting the attack (to the extent this is possible) is an
important objective.  While attack detection is a clear goal, the approach developed here does not necessarily offer any new insights for these
more general (and {\it quite} challenging) AD problems, except to the extent that
anomalous signatures for such problems might still manifest in a class-specific fashion.
\bibliographystyle{plain}
\bibliography{adversarial}
\end{document}